\documentclass{article} 

\let\CryoCueAddContentsLine\addcontentsline
\usepackage{iclr2027_conference,times}
\let\addcontentsline\CryoCueAddContentsLine

\iclrfinalcopy
\usepackage{etoolbox}
\makeatletter
\patchcmd{\@maketitle}
  {\lhead{Published as a conference paper at ICLR 2027}}{}
  {}{\PackageError{CryoCue}{Could not remove conference header}{Check the original style.}}
\patchcmd{\@maketitle}
  {\begin{tabular}[t]{l}\bf\rule{\z@}{24pt}\@author}
  {\begin{tabular}[t]{l}\bf\rule{\z@}{14pt}\@author}
  {}{\PackageError{CryoCue}{Could not adjust author spacing}{Check the original style.}}
\patchcmd{\@maketitle}{\vskip 0.3in minus 0.1in}{\vskip 6pt}
  {}{\PackageError{CryoCue}{Could not adjust title spacing}{Check the original style.}}
\makeatother
\makeatletter
\@ifpackageloaded{natbib}{}{\usepackage{natbib}}
\makeatother

\usepackage{amsmath,amssymb,mathtools}
\usepackage{graphicx}
\usepackage{booktabs}
\usepackage{multirow}
\usepackage{array}
\usepackage{xcolor}
\usepackage{xspace}
\usepackage{microtype}
\usepackage{float}
\floatstyle{ruled}
\newfloat{algorithm}{tbp}{loa}
\floatname{algorithm}{Algorithm}
\floatstyle{plain}
\usepackage{algpseudocode}
\usepackage[section]{placeins}
\usepackage{url}
\usepackage[hidelinks]{hyperref}
\usepackage{iftex}
\ifPDFTeX\else
  \usepackage{fontspec}
\fi
\newcommand{\method}{\textsc{CryoCue}\xspace}
\newcommand{\angstrom}{\textup{\AA}}

\newcommand{\nallmaps}{18{,}420\xspace}
\newcommand{\ntrainmaps}{17{,}820\xspace}
\newcommand{\nvalmaps}{300\xspace}
\newcommand{\nbenchmarkmaps}{300\xspace}
\newcommand{\ntraincrops}{185{,}373{,}427\xspace}
\newcommand{\nvalcrops}{3{,}543{,}489\xspace}
\newcommand{\nbenchmarkcrops}{2{,}029{,}806\xspace}

\title{Learning from Hetero Density for Cryo-EM Protein Reconstruction}

\author{%
  Xu Han\textsuperscript{1,2,3}\quad
  Chaozhuo Li\textsuperscript{4}\quad
  Xiaowei Yuan\textsuperscript{5}\\
  \bfseries Yuancheng Sun\textsuperscript{3}\quad
  Kang Liu\textsuperscript{1,2,3}\quad
  Qiwei Ye\textsuperscript{3,*}\\[4pt]
  \parbox[t]{\dimexpr\textwidth-2\tabcolsep\relax}{\normalfont\small\raggedright
    \textsuperscript{1}University of Chinese Academy of Sciences, Beijing, China\\
    \textsuperscript{2}Key Laboratory of Complex Systems Cognition and Decision,\\
    Institute of Automation, Chinese Academy of Sciences, Beijing, China\\
    \textsuperscript{3}Beijing Academy of Artificial Intelligence, Beijing, China\\
    \textsuperscript{4}Beijing University of Posts and Telecommunications, Beijing, China\\
    \textsuperscript{5}Ant Group, China\\[3pt]
    \textsuperscript{*}Corresponding author: \href{mailto:qiwei.ye@baai.ac.cn}{\texttt{qiwei.ye@baai.ac.cn}}
  }%
}
\date{}

\begin{document}
\maketitle

\begin{abstract}

Reconstructing protein structures from cryo-electron microscopy (cryo-EM) maps is essential for understanding macromolecular assemblies. Although learning-based methods have improved protein reconstruction, information from hetero components remains underused. Our analysis finds both false predictions and reference protein sites near hetero components; filtering nearby candidates can improve or impair chain construction. We introduce \method, a framework that uses hetero information to guide protein reconstruction. An anchor-supervised detector learns hetero representations across five component classes. Multiscale hetero features guide backbone localization, while predicted hetero candidates condition structure refinement through their class, confidence, and frame-relative geometry. Experiments show that \method improves backbone localization near hetero components and achieves more accurate protein structure reconstruction.

\end{abstract}

\section{Introduction}
\label{sec:introduction}

Cryo-electron microscopy (cryo-EM) is an important technique for
determining the structures of proteins and protein complexes
\citep{wwpdb2024emdb}.
The resulting three-dimensional density maps are interpreted
to reconstruct protein structures.
The general paradigm consists of backbone localization in the
maps and structure modeling from the predicted atom positions
\citep{terashi2024deepmainmast}.
Accurate protein structures reveal subunit interactions,
ligand-binding sites, and distinct conformational states.
These structural details help explain protein function and guide
structure-based drug design
\citep{su2026cryoatom,varga2025structure}.

Learning-based methods have improved protein reconstruction from
cryo-EM maps.
DeepTracer and Cryo2Struct predict backbone atom locations and
residue types from density
\citep{pfab2021deeptracer,giri2024novo}.
To guide residue assignment, ModelAngelo and EModelX incorporate
protein sequences \citep{jamali2024automated,chen2024protein}.
CryoAtom and EMProt use learned protein geometry to model
relationships between residues \citep{su2026cryoatom,li2026emprot}.
These methods combine density evidence with protein-specific
knowledge to improve reconstruction accuracy.

\begin{figure}[t]
    \centering
    \includegraphics[width=\linewidth]{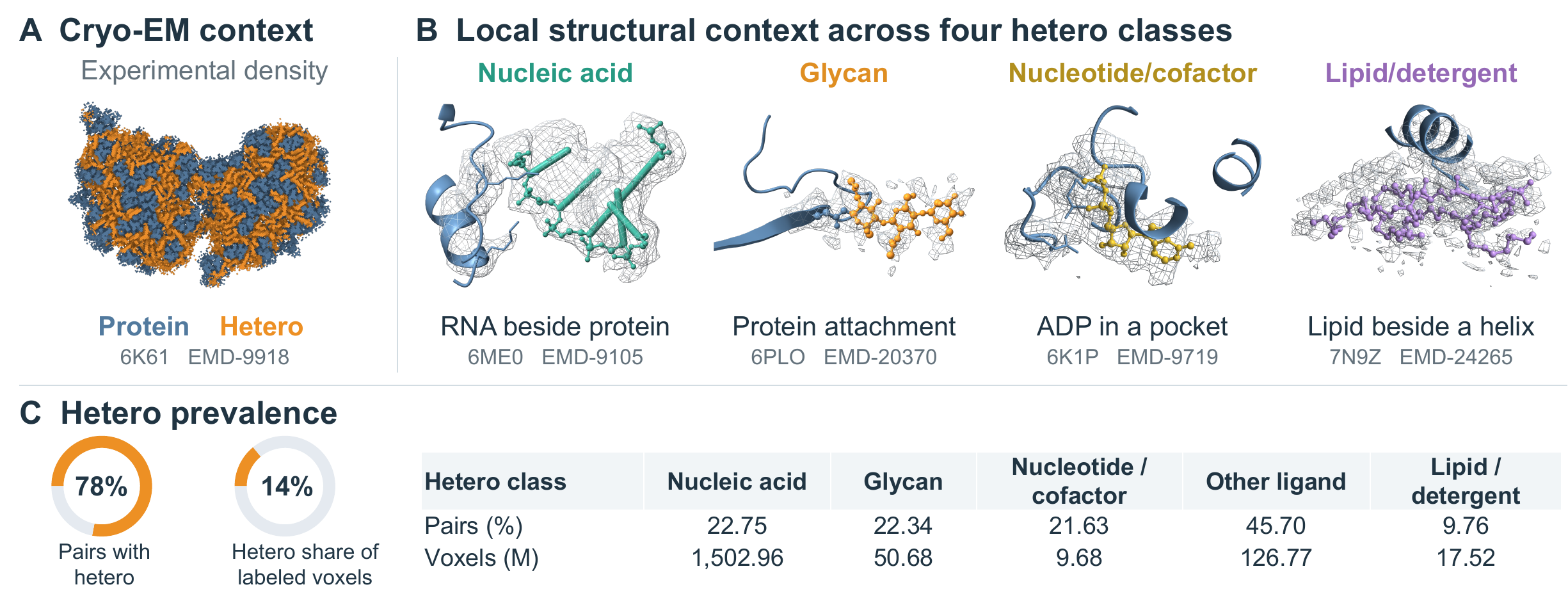}
    \caption{Hetero context in cryo-EM maps.
    \textbf{A,} Density assigned to protein and hetero using reference coordinates.
    \textbf{B,} Local contacts between protein and hetero components in reference structures;
    gray meshes show experimental density.
    \textbf{C,} Hetero prevalence and voxel counts across the sample dataset.
    M denotes million voxels.}
    \label{fig:hetero_context}
\end{figure}

Training data remain limited because building reliable reference
structures requires substantial time and expertise
\citep{jamali2024automated}.
EMDB contains about 62,000 maps, but only a subset have
corresponding structures
\citep{wwpdb2024emdb}.
Supervision for protein reconstruction mainly focuses on protein
annotations \citep{giri2024cryo2structdata}, although the same maps
also contain nucleic acids, glycans, cofactors, ligands, and
lipids or detergents.
We refer to these non-protein molecules as \emph{hetero components}
and their map signal as \emph{hetero density}
\citep{caspy2021dimeric}.
Figure~\ref{fig:hetero_context}A shows protein and hetero density
within the same assembly~\citep{zheng2019structural}, while Fig.~\ref{fig:hetero_context}B 
illustrates contacts between proteins and several hetero classes.
Together with the statistics in Fig.~\ref{fig:hetero_context}C,
these examples show that hetero components are widespread
across the analyzed maps.

The prevalence of hetero components raises an underexplored
question: can hetero information be used to improve protein
reconstruction?
Hetero density can resemble protein density and create ambiguity during
protein reconstruction.
ModelAngelo, for example, learns to ignore cofactor density rather
than build protein residues in these regions
\citep{jamali2024automated}.
However, hetero components can also help interpret neighboring
protein sites.
N-linked glycans are covalently attached to asparagine side
chains, linking their positions to specific protein sites
\citep{emsley2018structural}.
Figure~\ref{fig:hetero_context}B illustrates this attachment
\citep{yu2021mechanism}, alongside an interface between protein and
RNA, an ADP binding pocket, and a lipid adjacent to a protein
helix \citep{haack2019cryo,yan2019structures,li2021cryo}.
Existing work on nucleic acid reconstruction and component
modeling shows that non-protein density can be interpreted
computationally \citep{wang2023cryoread,mostosi2020haruspex}.
These observations suggest that hetero components can both
complicate protein localization and provide useful structural
clues.

To answer this question, we conduct a preliminary analysis of how
hetero components affect backbone localization and chain construction
(Section~\ref{sec:characterization}).
A straightforward strategy is to treat hetero neighborhoods uniformly,
for example by suppressing nearby candidate protein sites.
However, the effect is mixed: filtering improves some cases but
degrades others, even within the same hetero class.
Our analysis shows that some hetero neighborhoods are dominated by
false predictions, whereas others also contain nearby reference
protein sites.
This variation reflects the different structural contexts in which
hetero components interact with proteins.
These findings motivate using hetero information selectively,
according to component type and local protein context.

Using hetero information presents two challenges.
First, hetero components vary substantially in scale and geometry:
nucleic acids form extended polymers, whereas ligands and membrane
lipids can appear as compact bound molecules or elongated structures
\citep{wang2023cryoread,lawson2024outcomes,vsolinc2025cryo}.
Their density patterns therefore differ considerably across component
classes and can lie close to protein density.
Second, hetero information is not uniformly useful for protein
reconstruction.
Depending on the component class and local protein context, a hetero
neighborhood may contain false protein predictions, reference protein
sites, or both.
The key challenge is therefore to exploit informative hetero cues
without allowing misleading ones to interfere with protein reconstruction.

We propose \method, a novel framework for cryo-EM protein
reconstruction that adaptively exploits hetero information.
\method consists of two hetero-guided stages.
First, an anchor-supervised detector represents diverse hetero
components through representative sites and multiscale features.
These multiscale features guide backbone localization despite
the diverse scales and geometries of hetero components.
Second, the predicted protein sites initialize structure refinement,
where nearby hetero candidates are incorporated according to their
class, confidence, and relative geometry.
This adaptive conditioning allows the hetero contribution to vary as the
protein structure evolves.
Experiments show that \method improves both backbone localization
and final protein reconstruction, including hetero-associated regions.

Our contributions are:
\begin{itemize}
    \item We formulate and study the rarely explored problem of exploiting 
    hetero information for cryo-EM protein reconstruction.
    Our analysis shows that hetero neighborhoods can contain both
    false protein predictions and nearby reference protein sites, so
    treating them uniformly can either help or harm reconstruction.

\begin{figure}[t]
\centering
\includegraphics[width=\linewidth]{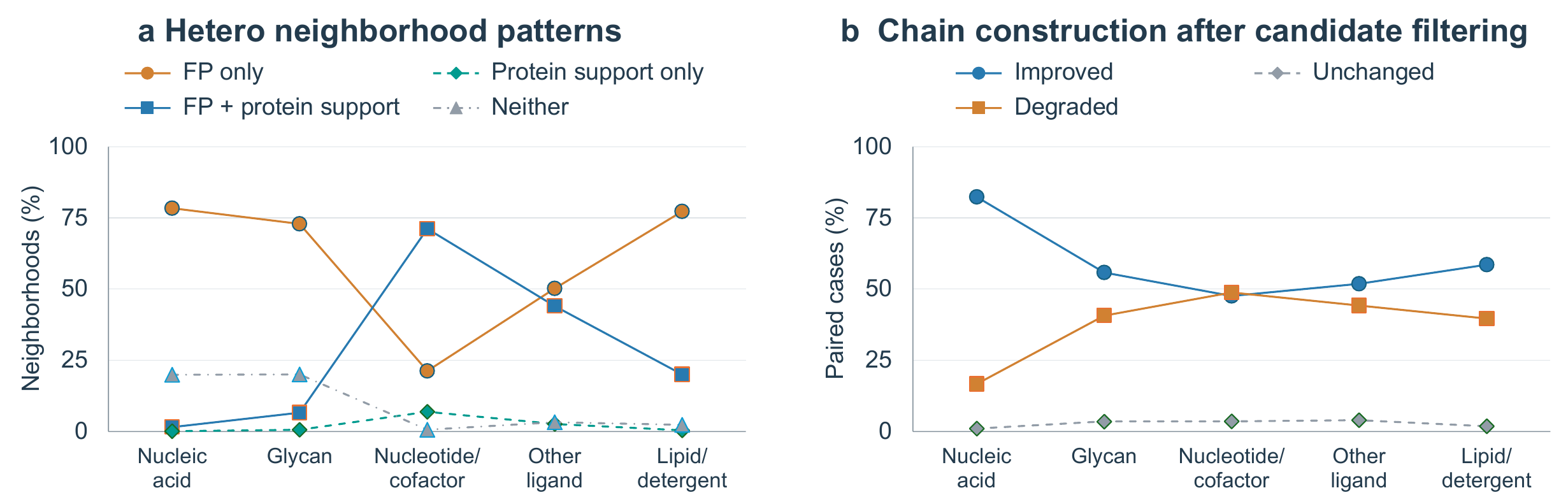}
\caption{Analysis of protein candidates near hetero components.
\textbf{a,} Patterns of false protein candidates and protein support
near hetero components.
\textbf{b,} Effects of candidate filtering on chain construction,
shown as improved, degraded, or unchanged.}
\label{fig:hetero_analysis}
\end{figure}

    \item We introduce \method to address two key challenges: representing
    hetero components with diverse scales and geometries, and using
    hetero information selectively without disrupting protein reconstruction.
    The framework combines a unified hetero representation with adaptive
    conditioning throughout backbone localization and structure refinement.

    \item Experiments show that \method improves both backbone localization
    and final protein reconstruction over protein-only baselines.
    The gains also extend to regions associated with hetero components.
\end{itemize}

\section{Preliminary Analysis}
\label{sec:characterization}

Cryo-EM protein reconstruction generally consists of backbone localization
followed by chain construction.
To examine whether hetero components affect both phases, we conduct a
preliminary analysis using Cryo2Struct, whose explicit candidate stage
allows predicted protein sites to be inspected before chain construction
\citep{giri2024novo}.
We first examine protein candidates near hetero components to understand
their effects on backbone localization.
Then, we test whether filtering candidates in these regions improves or
degrades chain construction.
Appendix~\ref{app:audit} gives the sampling procedure, protocols, and
statistics.

\paragraph{Hetero effects on backbone localization.}
In the first stage of protein reconstruction, hetero components can either
confuse or inform backbone localization.
We examine protein candidates near annotated hetero components and their
relationship to nearby protein support (Fig.~\ref{fig:hetero_analysis}a).
Nucleic acids, glycans, and lipids or detergents are often associated with
false candidates without protein support, indicating that their density can
be mistaken for protein.
In contrast, near nucleotides or cofactors, false candidates frequently
coexist with protein support, while other ligands show a mixture of these
patterns.
These differences across hetero classes show that hetero density should not
be treated as a uniform non-protein signal during backbone localization.
Its contribution can instead vary with the hetero class and local protein
context.

\paragraph{Hetero effects on chain construction.}
These mixed neighborhood patterns motivate us to test whether hetero
components also affect chain construction.
To test this effect, we filter protein candidates near annotated hetero
components before chain construction while keeping the remaining inputs
unchanged.
Filtering often improves the reconstructed chains, particularly around
nucleic acids, but degradations occur across all hetero classes
(Fig.~\ref{fig:hetero_analysis}b).
These mixed outcomes show that candidate filtering can remove misleading
protein predictions but can also discard useful protein information.
Together, these results show that hetero information has mixed effects
across both stages, motivating different treatment across hetero classes
based on nearby protein evidence.

\section{Methodology}
\label{sec:method}

\begin{figure}[t]
    \centering
    \includegraphics[width=\linewidth]{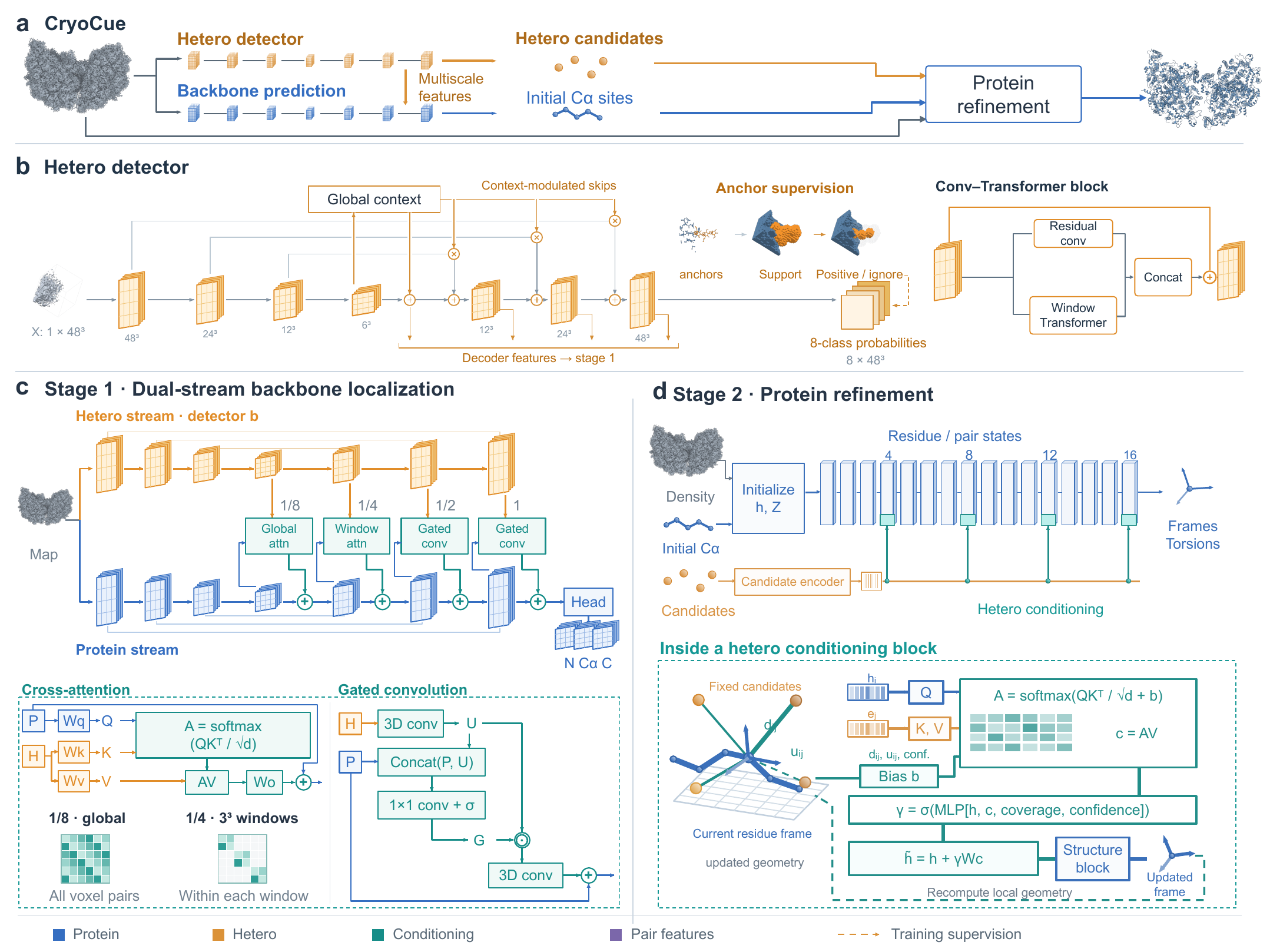}
    \caption{\textbf{Overview of \method.}
    \textbf{a,} Hetero-guided protein reconstruction.
    \textbf{b,} Anchor-supervised detection produces multiscale features and class predictions.
    \textbf{c,} Coarse attention and fine-scale gating condition backbone localization.
    \textbf{d,} Predicted hetero candidates condition refinement through their
    class, confidence, and geometry relative to evolving residue frames.}
    \label{fig:model_overview}
\end{figure}

Given a cryo-EM map and protein sequences, protein reconstruction proceeds
in two stages: backbone localization followed by structure refinement
(Fig.~\ref{fig:model_overview}).
\method incorporates hetero information into both stages.
An anchor-supervised detector provides multiscale hetero features for
backbone localization and sparse hetero candidates for structure refinement.
The first stage predicts initial protein sites through multiscale feature
fusion, while the second refines these sites using frame-relative hetero
conditioning.
Protein sequences are introduced only during the final chain assembly.

\subsection{Anchor-Supervised Hetero Representation}
\label{sec:hetero_detection}

Hetero components vary substantially in size and shape, making uniform
supervision difficult.
We therefore use representative reference atoms as anchors to define
localized targets (Fig.~\ref{fig:model_overview}b).
Voxels near an anchor are labeled positive only when their class matches
the anchor, while other annotated hetero regions are ignored.
This provides a consistent supervision target across diverse component
geometries.
Detailed anchor definitions and masking rules are given in
Appendix~\ref{app:anchor_supervision}.

The hetero detector uses a 3D adaptation of SCUNet
\citep{zhang2023practical}.
Its local blocks combine convolutional features with window-based Transformer
features \citep{liu2021swin} to capture local density structure and surrounding
context.
At the coarsest scale, full self-attention further provides global context.

The prediction head distinguishes background, protein backbone, protein
side chain, and five hetero classes: polymeric nucleic acid, glycan,
nucleotide/cofactor, other ligand, and lipid/detergent.
Including the protein classes provides explicit supervision for
distinguishing hetero from nearby protein density.
The detector is trained with focal cross-entropy and foreground Dice loss
\citep{lin2017focal,milletari2016v},
\begin{equation}
\mathcal L_h
=
\mathcal L_{\mathrm{focal}}
+
0.5\,\mathcal L_{\mathrm{Dice}}.
\label{eq:hetero_detection_loss}
\end{equation}
Ignored voxels are excluded from both terms.
Its multiscale features guide backbone localization, while predicted class
probabilities are used to extract hetero candidates for structure
refinement.
Further architecture details are provided in
Appendix~\ref{app:hetero_detector}.

\subsection{Backbone Localization with Hetero Features}
\label{sec:backbone_prediction}

The backbone predictor uses the same local 3D SCUNet architecture as the
hetero detector with independent parameters.
During localization, the two networks form parallel protein and hetero
streams: the protein stream predicts backbone responses, while the fixed
hetero stream provides features at matching resolutions.
Coarse hetero features provide broader context for separating protein and
hetero density, whereas fine features preserve local details for locating
protein sites.
We therefore use cross-attention at coarse resolutions and gated fusion at
fine resolutions.
At resolution $s$, protein feature $F_s^p$ receives a hetero update
$\Delta_s$.

\paragraph{Coarse context.}
At coarse resolutions, protein features query the corresponding hetero
features:
\begin{equation}
\Delta_s=W_s^O\operatorname{Attn}_s\!\left(
Q_s(F_s^p),K_s(F_s^h),V_s(F_s^h)\right),
\qquad s\in\{8,4\}.
\label{eq:backbone_attention_main}
\end{equation}
Protein features provide queries, while hetero features provide keys and
values.
Attention covers the full crop at the coarsest resolution and local
windows at the next resolution.
This allows each protein location to use hetero information beyond the
same spatial position.

\paragraph{Fine-scale fusion.}
At finer resolutions, protein and hetero features are spatially aligned.
We therefore use a gate to control the local hetero contribution:
\begin{equation}
\begin{aligned}
U_s&=\phi_s(F_s^h),\\
G_s&=\sigma\!\left(g_s([F_s^p;U_s])\right),\\
\Delta_s&=\psi_s(G_s\odot U_s),
\qquad s\in\{2,1\}.
\end{aligned}
\label{eq:backbone_gating_main}
\end{equation}
Here, $\phi_s$ and $\psi_s$ are convolutional projections, $g_s$
predicts the gate, and $[\,;\,]$, $\sigma$, and $\odot$ denote
concatenation, sigmoid, and elementwise multiplication.
Because the gate depends on both protein and hetero features, the hetero
contribution can vary across spatial locations.

We train the backbone predictor while keeping the hetero detector fixed.
Because background voxels greatly outnumber regions near backbone atoms,
we balance the regression loss between the two groups for each backbone
response:
\begin{equation}
\mathcal L_{\mathrm{voxel}}
=
\frac{1}{3}
\sum_a
\sum_{q\in\{+,-\}}
\frac{1}{2|\mathcal V_a^q|}
\sum_{v\in\mathcal V_a^q}
\left|
\frac{\widehat A_a(v)-A_a(v)}{100}
\right|.
\label{eq:backbone_loss_main}
\end{equation}
Here, $a$ indexes the three backbone response channels, and
$\mathcal V_a^+$ and $\mathcal V_a^-$ denote atom-near and background
voxels, respectively.
The resulting predictions define initial protein sites for structure
refinement, while a separate amino acid predictor provides residue-type
scores for later sequence assignment.
Detailed architecture and fusion settings are given in
Appendix~\ref{app:backbone_architecture}.

\subsection{Structure Refinement with Hetero Candidates}
\label{sec:protein_reconstruction}

The second stage refines the initial protein sites into a protein structure.
Its backbone is a three-track structure network with invariant point
attention (IPA), jointly updating residue features, pair features, and
local protein frames
\citep{baek2021accurate,jumper2021highly,li2026emprot}.
We inject hetero conditioning at selected structure blocks.
As protein frames move and rotate during refinement, detected hetero
candidates remain fixed in the map.
We therefore represent each nearby candidate relative to the current
protein frame and use this geometry to update protein features.

\paragraph{Candidate representation.}
Each hetero candidate has a fixed position, predicted class probabilities,
and confidence.
A shared encoder maps its class probabilities, confidence, and prediction
uncertainty to an embedding $e_j$.
For each initial protein site, we select a fixed neighborhood
$\mathcal N_i$ of nearby candidates, whose positions and membership remain
unchanged during refinement.

\paragraph{Frame-relative geometry.}
As a protein frame moves, its relation to the fixed hetero candidates
changes.
For protein frame $T_i=(R_i,t_i)$ and candidate position $y_j$, we compute
\begin{equation}
d_{ij}=\|y_j-t_i\|_2,
\qquad
u_{ij}=
\frac{R_i^\top(y_j-t_i)}
{\max(d_{ij},\epsilon)}.
\label{eq:hetero_geometry_main}
\end{equation}
The distance $d_{ij}$ and local direction $u_{ij}$ are recomputed from the
current frame at every conditioning step, keeping the hetero geometry
aligned with the evolving protein structure.

\paragraph{Hetero conditioning.}
At each conditioning step, protein features query nearby hetero candidates,
whose embeddings provide keys and values.
For one attention head,
\begin{equation}
\begin{aligned}
\alpha_{ij}
&=
\operatorname{softmax}_{j\in\mathcal N_i}
\left(
\frac{q_i^\top k_j}{\sqrt{d_h}}
+
b(d_{ij},u_{ij},o_j)
\right),\\
c_i
&=
\sum_{j\in\mathcal N_i}\alpha_{ij}v_j .
\end{aligned}
\label{eq:hetero_conditioning_main}
\end{equation}
The attention bias incorporates frame-relative geometry and candidate
confidence, while the candidate embedding carries class information.
After concatenating the attention heads, the resulting context
$c_i^{\mathrm{all}}$ updates each protein feature through a learned gate:
\begin{equation}
\widetilde h_i
=
h_i+\gamma_iP_O(c_i^{\mathrm{all}}),
\qquad
(H',Z',T')
=
\mathcal S(\widetilde H,Z,T).
\label{eq:hetero_frame_update_main}
\end{equation}
Thus, hetero information contributes adaptively at each protein site.
The candidate neighborhood remains fixed, while its geometry is recomputed
as the protein structure evolves.

The refined frames and residue features are used to predict backbone
geometry and torsions.
They first define provisional chain fragments.
Residue-type scores sampled at the refined protein sites form profiles for
alignment to the input protein sequences.
The resulting sequence assignments guide fragment assembly and final
all-atom construction.

We train the structure network with backbone-coordinate and torsion
supervision:
\begin{equation}
\begin{aligned}
\mathcal L_{\mathrm{backbone}}
&=
\operatorname*{mean}_{(i,a)\in\mathcal M_{\mathrm{bb}}}
\rho\!\left(
\|\widehat x_{ia}-x_{ia}\|_2
\right),\\
\mathcal L_{\mathrm{torsion}}
&=
\operatorname*{mean}_{(i,k)\in\mathcal M_{\mathrm{tor}}}
\min\!\left(
\|\widehat z_{ik}-z_{ik}\|_2^2,
\|\widehat z_{ik}-z_{ik}^{\mathrm{alt}}\|_2^2
\right),\\
\mathcal L_{\mathrm{structure}}
&=
\mathcal L_{\mathrm{backbone}}
+
0.1\,\mathcal L_{\mathrm{torsion}}.
\end{aligned}
\label{eq:structure_loss_main}
\end{equation}
Here, $\rho$ denotes the radial Huber loss for backbone coordinates,
while $\widehat z_{ik}$ is the normalized predicted torsion pair and
$z_{ik}^{\mathrm{alt}}$ accounts for equivalent torsion representations.
Appendix~\ref{app:protein_architecture} provides further details on the
structure architecture, candidate encoding, loss definitions, and final assembly.

\section{Experiments}
\label{sec:experiments}
\label{sec:results}

\begin{table}[t]
  \centering
    \caption{Protein reconstruction performance.
    Results on the test set.
    Values are map-level means; bold indicates the best mean.}
  \label{tab:structure_keep300}
  \begingroup
  \small
  \setlength{\tabcolsep}{4pt}
  \renewcommand{\arraystretch}{1.15}
  \resizebox{\linewidth}{!}{%
  \begin{tabular}{@{}lrrrrrr@{}}
    \toprule
    & \multicolumn{4}{c}{Structure reconstruction}
    & \multicolumn{1}{c}{Density support}
    & \multicolumn{1}{c}{Connectivity} \\
    \cmidrule(lr){2-5}\cmidrule(lr){6-6}\cmidrule(l){7-7}
    Method
    & TM-score $\uparrow$
    & lDDT $\uparrow$
    & \shortstack{Coverage (\%) $\uparrow$}
    & \shortstack{RMSD (\AA) $\downarrow$}
    & Q-score $\uparrow$
    & \shortstack{Break rate (\%) $\downarrow$} \\
    \midrule
    EMProt & 0.805 & 0.740 & 81.07 & 0.576 & 0.604 & 1.20 \\
    E3-CryoFold & 0.711 & 0.512 & 67.05 & 1.085 & 0.369 & 8.72 \\
    ModelAngelo & 0.728 & 0.658 & 73.22 & 0.592 & \textbf{0.613} & 2.61 \\
    EModelX & 0.805 & 0.687 & 82.43 & 0.842 & 0.447 & 6.96 \\
    \midrule
    \textsc{CryoCue} (ours)
    & \textbf{0.846}
    & \textbf{0.789}
    & \textbf{85.02}
    & \textbf{0.571}
    & 0.593
    & \textbf{0.69} \\
    \bottomrule
  \end{tabular}%
  }
  \endgroup
  \par\vspace{-6pt}
\end{table}

\begin{table}[th]
  \centering
  \caption{Backbone localization and protein--hetero contact.
  All scores are percentages; higher is better.
  Bold indicates the best mean among the compared methods.}
  \label{tab:hetero_keep300}
  \begingroup
  \small
  \setlength{\tabcolsep}{5pt}
  \renewcommand{\arraystretch}{1.15}
  \resizebox{\linewidth}{!}{%
  \begin{tabular}{@{}lrrrrrr@{}}
    \toprule
    & \multicolumn{2}{c}{Overall backbone}
    & \multicolumn{2}{c}{Hetero neighborhood}
    & \multicolumn{2}{c}{Protein--hetero contacts} \\
    \cmidrule(lr){2-3}\cmidrule(lr){4-5}\cmidrule(l){6-7}
    Method
    & CA precision $\uparrow$ & CA recall $\uparrow$
    & CA precision $\uparrow$ & CA recall $\uparrow$
    & Precision $\uparrow$ & Recall $\uparrow$ \\
    \midrule
    EMProt & 96.534 & 82.675 & 93.035 & 86.181 & 85.646 & 73.993 \\
    E3-CryoFold & 86.013 & 79.818 & 71.638 & 76.606 & 67.762 & 59.699 \\
    ModelAngelo & 93.688 & 81.225 & \textbf{97.511} & 84.937 & \textbf{87.765} & 73.145 \\
    EModelX & 92.946 & 83.557 & 75.846 & 86.832 & 71.419 & 64.636 \\
    \midrule
    \textsc{CryoCue} (ours) & \textbf{96.897} & \textbf{85.693} & 96.271 & \textbf{89.883} & 86.396 & \textbf{77.877} \\
    \bottomrule
  \end{tabular}%
  }
  \endgroup
  \par\vspace{-6pt}
\end{table}

\subsection{Experimental Setup}
\label{sec:experimental_setup}

\paragraph{Datasets and preprocessing.}
Following prior cryo-EM reconstruction work
\citep{li2026emprot,giri2024novo},
we collect single-particle map--structure pairs containing protein chains.
We construct family-aware training, validation, and test splits and remove
sequence-similar training pairs using MMseqs2.
Maps are resampled, normalized, and processed as local density crops.
Appendix~\ref{app:dataset-construction} provides the complete dataset,
preprocessing, and supervision protocol, while
Appendix~\ref{app:training} describes training and optimization.

\paragraph{Baselines and evaluation metrics.}
We compare \method with EMProt \citep{li2026emprot},
E3-CryoFold \citep{wang2025end},
ModelAngelo \citep{jamali2024automated}, and EModelX
\citep{chen2024protein}, using the inputs supported by each method.
We evaluate global structure reconstruction, local geometry, density support,
backbone localization, and recovery near hetero components using TM-score,
Local Distance Difference Test (lDDT), coverage, Q-score, backbone precision, recall, and RMSD, together with
hetero-local backbone and protein--hetero contact metrics.
Appendix~\ref{app:evaluation_metrics} provides the complete metric
definitions, matching procedures, and aggregation protocol.

\subsection{Protein Structure Reconstruction}
\label{sec:global_results}

We first evaluate the reconstructed protein structures in
Table~\ref{tab:structure_keep300}.
TM-score measures global structural agreement, lDDT reflects local
backbone geometry, and coverage measures how much of the reference
structure is recovered.
\method improves all three metrics, showing that the reconstructed
structures are both more accurate and more complete.

The improvement also extends to chain continuity.
The lower break rate indicates fewer missing connections between recovered
protein sites, suggesting that the gain is not limited to localization but
also leads to more continuous protein chains.
This is consistent with our preliminary analysis, where hetero neighborhoods
were shown to affect chain construction.

RMSD and Q-score show the boundary of this improvement.
Because RMSD is computed only over successfully matched backbone positions,
the smaller change suggests that the main benefit lies in recovering and
organizing protein regions.
The effect is more limited for coordinates that are already matched.
Q-score does not improve in the same way, indicating that better structural
reconstruction does not necessarily imply better density fit.
Overall, hetero information mainly improves structure recovery and chain
continuity, rather than producing large changes in matched-coordinate
accuracy or density support.

\subsection{Backbone Localization near Hetero Components}
\label{sec:local_results}

We next examine backbone localization, with particular attention to regions
near hetero components (Table~\ref{tab:hetero_keep300}).
Precision reflects false protein predictions, while recall measures the
recovery of reference protein sites.
\method achieves the highest overall precision and recall, showing that the
additional recovery is not obtained by simply producing more predictions.

\begin{table}[t]
\centering
\caption{Ablation of hetero representation and conditioning.
Variants isolate hetero fusion during backbone localization, anchor-based
supervision, and dynamic frame-relative conditioning.}
\label{tab:ablation}
\begingroup
\small
\setlength{\tabcolsep}{3.2pt}
\renewcommand{\arraystretch}{1.12}
\resizebox{\linewidth}{!}{%
\begin{tabular}{@{}lrrrrrr@{}}
\toprule
& \multicolumn{4}{c}{Structure reconstruction}
& \multicolumn{1}{c}{Density}
& \multicolumn{1}{c}{Connectivity} \\
\cmidrule(lr){2-5}
\cmidrule(lr){6-6}
\cmidrule(l){7-7}
Method
& TM-score $\uparrow$
& lDDT $\uparrow$
& \shortstack{Coverage (\%) $\uparrow$}
& \shortstack{RMSD (\AA) $\downarrow$}
& Q-score $\uparrow$
& \shortstack{Break rate (\%) $\downarrow$} \\
\midrule
Protein-only & 0.780 & 0.711 & 77.44 & 0.609 & 0.602 & 1.54 \\
Backbone fusion only
 & 0.819 & 0.760 & 82.53 & 0.588 & \textbf{0.622} & 1.06 \\
Full w/o anchor supervision & 0.747 & 0.692 & 75.80 & 0.715 & 0.545 & 1.78 \\
Full w/ fixed hetero geometry & 0.840 & 0.782 & 84.54 & 0.573 & 0.596 & 0.75 \\
Full model
& \textbf{0.846}
& \textbf{0.789}
& \textbf{85.02}
& \textbf{0.571}
& 0.593
& \textbf{0.69} \\

\bottomrule
\end{tabular}%
}
\endgroup
\end{table}

\begin{figure}[th]
\centering
\includegraphics[width=\linewidth]{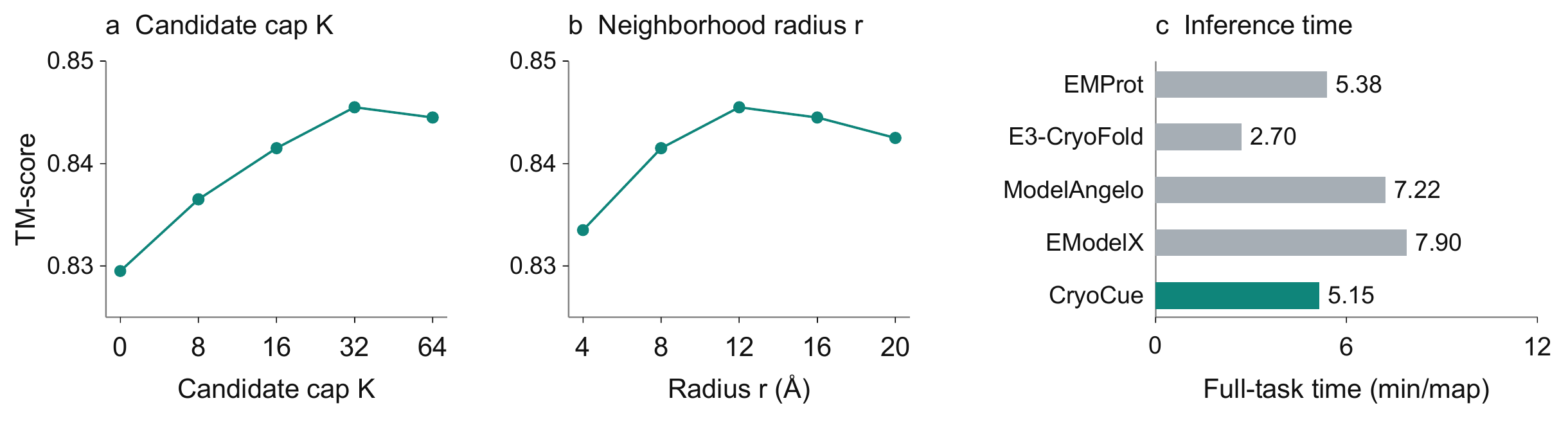}
\caption{Hyperparameter sensitivity and inference efficiency.
Left: TM-score under different candidate caps $K$ and neighborhood radii $r$,
with the other hyperparameter fixed.
Right: end-to-end inference time across methods.}
\label{fig:sensitivity}\label{fig:efficiency}
\end{figure}

The advantage becomes more apparent near hetero components.
\method achieves the highest hetero-neighborhood recall, while maintaining
high precision, indicating that more reference protein sites are recovered
in these regions.
The same trend appears in protein--hetero contacts: contact recall improves,
while precision remains close to the strongest baseline.
Together, these results show that the main gain lies in more complete
recovery of hetero-associated protein structure.

Detailed hetero detection results are provided in
Appendix~\ref{app:hetero_detection_results}.
Detection performance varies across component classes, with nucleic acids
detected more reliably than several smaller or more diverse classes.
Despite this variation, the predicted hetero context improves protein
recovery near hetero components.
Together with our preliminary analysis, these results suggest that hetero
context remains useful despite uneven detection quality across component
types.

\subsection{Ablation of Hetero Representation and Conditioning}
\label{sec:ablation}

We ablate the main components of hetero conditioning in
Table~\ref{tab:ablation}.
All hetero information is predicted from the cryo-EM map, while protein
sequences are used only during final chain construction and remain unchanged
across variants.
Using hetero features only during backbone localization already improves
structure reconstruction and chain continuity over the protein-only model.
Adding hetero conditioning during structure refinement brings further gains,
showing that hetero information contributes at both stages.

Importantly, simply expanding hetero supervision is not always beneficial.
Removing anchor supervision degrades the structural metrics even below the
protein-only variant.
This result agrees with our preliminary analysis: hetero regions can contain
both useful protein context and misleading signals.
Anchor supervision is therefore important for localizing informative hetero
cues rather than treating all hetero density as equally useful.

Dynamic geometry provides an additional gain during structure refinement.
Using fixed hetero geometry already performs strongly, but recomputing the
frame-relative geometry as the protein structure evolves consistently
improves reconstruction.
This supports modeling the changing relation between fixed hetero candidates
and evolving protein frames.

Q-score shows a different, also informative trend.
The Backbone fusion only variant achieves the strongest density support, whereas the
full model performs best on structure reconstruction and connectivity.
Thus, better structural reconstruction does not necessarily require a higher
density-fit score.
The refinement stage mainly improves the recovery and organization of the
protein structure.
Overall, these ablations show that the benefit of hetero information depends
not simply on adding more hetero signal, but on how it is supervised and
conditioned on the protein.

\subsection{Hyperparameter Sensitivity and Efficiency}
\label{sec:sensitivity_efficiency}

We first examine sensitivity to the amount of local hetero context (Figure~\ref{fig:sensitivity}).
Increasing the candidate cap and neighborhood radius from small settings
improves reconstruction, but the gains saturate and slightly decline as
the context becomes larger.
This non-monotonic trend suggests that useful hetero cues are primarily
local and that more context is not always beneficial.
Performance remains stable around the selected settings, indicating limited
sensitivity to the exact hyperparameters.

We next compare end-to-end inference time with existing methods.
\method has a runtime close to EMProt and is faster than ModelAngelo and
EModelX.
Thus, the additional hetero detection and conditioning introduce only
moderate overhead while retaining practical inference efficiency.

\subsection{Case Study}
\label{sec:case_study}

\begin{figure}[t]
    \centering
    \includegraphics[width=\linewidth]{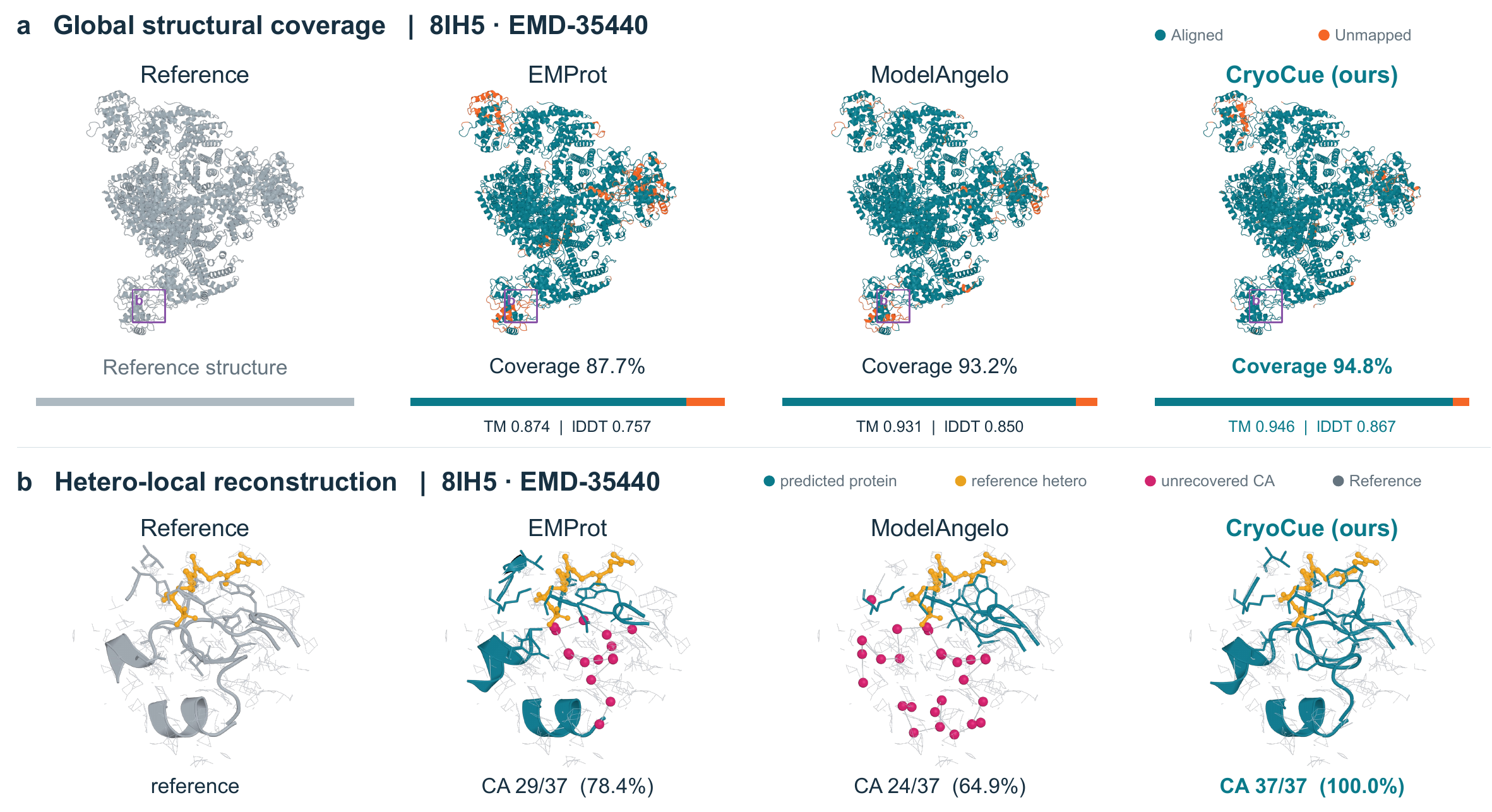}
    \caption{Case study of global reconstruction and recovery near a hetero component.
    \textbf{a,} Global reconstruction of 8IH5 (EMD-35440) by EMProt,
    ModelAngelo, and CryoCue; the highlighted region is enlarged below.
    \textbf{b,} Reconstruction near the corresponding hetero component.
    Magenta markers indicate unrecovered reference protein sites.}
    \label{fig:case_study}
    \vspace{-6pt}
\end{figure}

Figure~\ref{fig:case_study} illustrates how the reconstruction improvement
appears at both global and local scales.
Globally, \method recovers more of the reference structure while preserving
its overall organization.
In the highlighted hetero-associated region, the compared methods miss
several reference protein sites, whereas \method preserves the local protein
trace.
This example links improved recovery near a hetero component to the increase
in structural coverage at the whole-protein level.

\section{Conclusion}
\label{sec:conclusion}
\label{sec:discussion}

We show that hetero density can provide useful context for cryo-EM protein
reconstruction, but its contribution depends on the local structural context.
\method exploits this information during backbone localization and structure
refinement through anchor supervision and frame-relative conditioning.
The results show improved structure recovery and chain continuity.
The ablations further show the importance of anchor supervision and
dynamic hetero conditioning.

The current approach still depends on the quality of hetero predictions, and
its gains are clearer in structural recovery than in atom-level density fit.
Protein sequences are currently introduced only during final chain
construction.
Improving hetero prediction and using sequence information earlier are natural
directions for future work.

\section*{AI Use Statement}
Generative AI tools were used to assist with manuscript organization,
language editing, drafting, literature retrieval, research execution, and
software development.
This includes AI-assisted scripts and code used during the project.
All AI-assisted text and code were reviewed, revised, and tested by the
authors.
The authors take full responsibility for the final manuscript, code,
results, and scientific claims.

\section*{Ethics Statement}
This study uses publicly available cryo-EM maps and macromolecular
structures.
It does not involve human participants, private data, or personally
identifiable information.

\section*{Reproducibility Statement}
The appendix provides details on dataset construction, supervision,
model architectures, training, and evaluation protocols.

\bibliographystyle{iclr2027_conference}
\bibliography{references}

\appendix
\section{Related Work}
\label{sec:related}

\paragraph{Cryo-EM protein reconstruction.}
Learning-based methods increasingly combine density evidence with chain
construction, sequence information, and learned protein geometry.
DeepTracer predicts backbone and residue information directly from density,
while Cryo2Struct predicts backbone atom locations and amino acid types
before tracing protein chains with a hidden Markov model (HMM)
\citep{pfab2021deeptracer,giri2024novo}.
ModelAngelo and EModelX further incorporate protein sequences to guide
residue assignment \citep{jamali2024automated,chen2024protein}.
Other methods introduce stronger structural priors: DeepMainmast and
DiffModeler combine density-derived information with predicted protein
structures, E3-CryoFold uses an SE(3)-equivariant graph network, and
CryoAtom, EMProt, and CryoACE refine protein structures through learned
geometric representations or structure-generation networks
\citep{terashi2024deepmainmast,wang2024diffmodeler,
wang2025end,su2026cryoatom,li2026emprot,li2026cryoace}.
These methods primarily improve reconstruction using protein-derived
information. In contrast, we study whether hetero density can provide
additional evidence for backbone localization and structure refinement.

\paragraph{Non-protein density modeling.}
Existing methods model specific types of non-protein density, often with
component-specific priors.
Haruspex and Emap2sec+ distinguish nucleic acid density from protein,
while CryoREAD, EMRNA, and EM2NA reconstruct nucleic acid structures
\citep{mostosi2020haruspex,wang2021detecting,
wang2023cryoread,li2025all,li2024automated}.
GemSpot, ChemEM, EMERALD, and recent learned methods fit, detect, or
reconstruct small-molecule ligands
\citep{robertson2020gemspot,sweeney2024chemem,muenks2023automatic,li2026direct}.
LipIDens and SURFER address lipid or membrane-associated density
\citep{ansell2023lipidens,bharadwaj2026interactive}.
Rather than reconstructing individual hetero components as the primary
target, we use predicted hetero information to guide protein reconstruction.

\section{Dataset Details}
\label{app:dataset-construction}

\subsection{Data collection and quality control}

Following the data-selection criteria of EMProt, and
Cryo2Struct \citep{li2026emprot,giri2024novo},
we collect cryo-EM maps from EMDB \citep{wwpdb2024emdb} and fitted
structures from the RCSB PDB \citep{berman2000protein}. Each pair links
a released primary map to a fitted structure through an official
EMD--PDB association. We include pairs with and without annotated
hetero components, subject to the eligibility criteria and quality
checks in Table~\ref{tab:data-selection}.

\begin{table}[htbp]
\caption{Data collection and quality control. These checks determine
pair eligibility; local label validity is handled separately in
Appendix~\ref{app:supervision}.}
\label{tab:data-selection}
\centering
\small
\begin{tabular}{>{\raggedright\arraybackslash}p{0.21\linewidth}>{\raggedright\arraybackslash}p{0.71\linewidth}}
\toprule
Step & Criteria \\
\midrule
Database search &
Single-particle cryo-EM maps in EMDB and electron-microscopy
structures in RCSB, with reported resolution $\leq4$~\angstrom{}. \\
Pair eligibility &
A released primary map, an official fitted EMD--PDB association,
available map and coordinate files, and at least one protein chain. \\
Structure validation &
Parseable required mmCIF fields and reliable component classification
using the Chemical Component Dictionary (CCD)
\citep{westbrook2015chemical} and entity/polymer metadata. Pairs with unresolved classification conflicts are excluded. \\
Raw-map quality &
Complete downloads; MRC dimensions and voxel spacing consistent with
official metadata. \\
\bottomrule
\end{tabular}
\end{table}

We canonicalize EMD--PDB associations, consolidate repeated EMDB
entries across sources, and retain one eligible fitted structure
per map. We exclude unresolved duplicate associations and pairs
containing unknown or ambiguously classified components, rather
than assigning these components to the other-ligand class. Split
manifests record each selected map and fitted structure; all dataset
sizes below count these pairs.

\subsection{Dataset splits}

We first randomly sample validation and test pairs, using MMseqs2-based
protein-family assignments to maintain family diversity
\citep{steinegger2017mmseqs2}. We then compare every protein chain
in each remaining training candidate against all protein chains in the
validation and test sets. We exclude a candidate pair if any chain
alignment satisfies
\begin{equation}
\mathrm{identity}\geq0.25,
\qquad
\frac{\mathrm{alignment\ length}}
{\min(\mathrm{query\ length},\mathrm{target\ length})}\geq0.80.
\label{eq:dataset_homology}
\end{equation}
We use \texttt{MMseqs2 easy-search}, combining query-coverage and
target-coverage searches with minimum identity 0.25, coverage 0.8,
sensitivity 7.5, and at most 1,000,000 returned sequences. We also
exclude training candidates whose map or coordinate files are
identical to a validation or test file.

The resulting dataset contains \nallmaps pairs: \ntrainmaps for
training, \nvalmaps for validation, and \nbenchmarkmaps for testing
(Table~\ref{tab:dataset-summary}). We use validation data to select checkpoints and
hyperparameters.

\begin{table}[htbp]
\caption{Dataset split sizes and complete stride-16 crop-grid counts.
Grid counts include empty-density regions and boundary crops;
they are not counts of independently sampled training examples.}
\label{tab:dataset-summary}
\centering
\small
\begin{tabular}{lrr}
\toprule
Split or test subset & Pairs & Grid crops \\
\midrule
Training & \ntrainmaps & \ntraincrops \\
Validation & \nvalmaps & \nvalcrops \\
Test & \nbenchmarkmaps & \nbenchmarkcrops \\
\bottomrule
\end{tabular}
\end{table}

\subsection{Map preprocessing}

We place maps and fitted coordinates in a common physical frame
before generating labels (Table~\ref{tab:map-preprocessing}).
Map-level normalization ensures that overlapping crops contain
the same density values.

\begin{table}[htbp]
\caption{Map preprocessing for the detector and backbone network.}
\label{tab:map-preprocessing}
\centering
\small
\begin{tabular}{>{\raggedright\arraybackslash}p{0.21\linewidth}>{\raggedright\arraybackslash}p{0.71\linewidth}}
\toprule
Step & Operation \\
\midrule
Coordinate alignment &
Use MRC axis order, header origin, and grid starts to express
voxels and fitted coordinates in the same physical frame. \\
Resampling &
Resample to isotropic 1~\angstrom{} spacing using cubic
convolution ($a=-0.5$), preserving map--coordinate alignment. \\
Normalization &
Apply full-map quantile clipping and scaling, as defined below. \\
Input crops &
Extract $48^3$ crops at stride 16. Zero-pad positions outside
the map and exclude them from supervision. \\
\bottomrule
\end{tabular}
\end{table}

Let $q$ be the 99.999th percentile of the positive values in the
resampled map $V$. We normalize the full map once:
\begin{equation}
X=\frac{\operatorname{clip}(V,0,q)}{q+10^{-6}}.
\label{eq:dataset_map_normalization}
\end{equation}
We compute normalization in float32 and store processed maps in
float16, without renormalizing individual crops. The detector is
supervised on non-ignored voxels throughout the crop, and the
backbone predictor on the central $16^3$ region. Training sampling
is described in Appendix~\ref{app:training}.

\subsection{Label construction}
\label{app:supervision}

\paragraph{Structural classes.}
The labels distinguish hetero sites from nearby protein and provide
comparable supervision across component shapes. We derive them from
the first model of each fitted structure using the fixed taxonomy
applied during screening. We retain heavy atoms and resolve alternate
conformations by occupancy; the taxonomy specification gives the
complete atom-selection rules. The detector predicts background,
protein backbone (N, C$\alpha$, C, O, and OXT), protein side chain
(other retained protein atoms), and the five hetero classes in
Table~\ref{tab:classes-anchors}.

\begin{table}[htbp]
\caption{Hetero class definitions and supervision anchors. Named
anchors must be uniquely present in the corresponding residue or
component.}
\label{tab:classes-anchors}
\centering
\small
\begin{tabular}{>{\raggedright\arraybackslash}p{0.25\linewidth}>{\raggedright\arraybackslash}p{0.40\linewidth}>{\raggedright\arraybackslash}p{0.23\linewidth}}
\toprule
Class & Definition & Anchor \\
\midrule
Polymeric nucleic acid &
Nucleotides assigned to nucleic acid polymers &
C4$^\prime$ per nucleotide residue \\
Glycan &
Sugar residues, including NAG, MAN, and FUC &
C1 per sugar residue \\
Nucleotide/cofactor &
Free nucleotides and nucleotide-derived cofactors, including
NAD/NADH, FAD/FMN, and SAM/SAH &
C4$^\prime$ \\
Other ligand &
Other reliably classified non-polymer ligands &
Representative heavy atom \\
Lipid/detergent &
Classified lipid and detergent components &
Representative heavy atom \\
\bottomrule
\end{tabular}
\end{table}

\paragraph{Structural support.}
We assign each voxel to the class of its nearest retained heavy atom
within 3~\angstrom{}. We ignore assignments when the nearest distances
to two different classes differ by less than 0.25~\angstrom{}.
Unassigned zero-density voxels and water-associated regions receive
background labels; positive density without assigned structural
support is ignored. These rules govern supervision within retained
pairs and leave the input density unchanged.

\paragraph{Anchor-based supervision.}
\label{app:anchor_supervision}
Anchors specify where to localize each component, while structural
support determines which voxels can receive its label. We use named
anchors for nucleotides and sugars. For other ligands and
lipids/detergents, we select the retained heavy atom nearest the
component's heavy-atom centroid. For coordinates $\mathcal A_j$,
this anchor is
\begin{equation}
a_j=\operatorname*{arg\,min}_{a\in\mathcal A_j}
\sum_{x\in\mathcal A_j}\|a-x\|_2^2.
\label{eq:anchors}
\end{equation}
This criterion selects an actual atom, avoiding a centroid that may
lie outside the molecule. Ties follow a deterministic atom order.
Table~\ref{tab:classes-anchors} gives the other anchor definitions;
missing or non-unique named atoms produce no anchor.
Non-polymer components are eligible when their representative center
lies inside the map and has same-class structural support within
4~\angstrom{}. A hetero voxel is positive only if it lies within
3~\angstrom{} of an accepted anchor and its nearest anchor class
agrees with its structural-support class. Near-ties between anchor
classes use the same 0.25~\angstrom{} tolerance. We ignore hetero
support outside accepted anchor targets, including components without
an accepted anchor. This concentrates positive supervision around
representative sites without labeling the rest of an annotated hetero
component as background.

\paragraph{Backbone-atom targets.}
Targets have channel order $(N,\mathrm{CA},C)$. For atom type $a$
with reference coordinate set $\mathcal X_a$, the response is
\begin{equation}
A_a(v)=\max_{x\in\mathcal X_a}
100\exp\!\left[-\left(\frac{\pi}{4\,\angstrom{}}\right)^2
\|v-x\|_2^2\right],
\label{eq:protein_target_rendering}
\end{equation}
with values below 0.25 set to zero. Targets are rendered online
from sparse coordinates in the processed map frame.

\subsection{Dataset statistics}
\label{app:statistics}

Table~\ref{tab:dataset-class-presence} reports the number of pairs
whose fitted structures contain each hetero class; classes can
co-occur within a pair.

\begin{table}[htbp]
\caption{Pair-level presence of the five hetero classes. A pair is
counted once for each class present in its structural labels, so
counts across classes are not additive.}
\label{tab:dataset-class-presence}
\centering
\scriptsize
\resizebox{\linewidth}{!}{%
\begin{tabular}{lrrrrr}
\toprule
Split or test subset & Polymeric nucleic acid & Glycan & Nucleotide/cofactor & Other ligand & Lipid/detergent \\
\midrule
Training & 4,040 & 3,983 & 3,859 & 8,161 & 1,743 \\
Validation & 104 & 31 & 68 & 99 & 17 \\
Test & 110 & 32 & 53 & 98 & 21 \\
\bottomrule
\end{tabular}}
\end{table}

The masked structural-support census contains 484,570,312,002
training voxels (Table~\ref{tab:training-support-census}). Protein
backbone and side-chain support together occupy 1.8063\%, and the
five hetero classes occupy 0.3451\%. These counts describe structural
support before anchor restriction; they do not represent the
distribution of sampled training inputs. Appendix~\ref{app:audit}
reports full-map occupancy on the analysis subset.

\begin{table}[htbp]
\caption{Class composition of the masked structural-support census
in the training split, before anchor-target restriction.}
\label{tab:training-support-census}
\centering
\small
\begin{tabular}{lrr}
\toprule
Support class & Voxels & Share \\
\midrule
Background & 474,145,598,844 & 97.8487\% \\
Protein backbone & 3,256,567,203 & 0.6721\% \\
Protein side chain & 5,496,114,368 & 1.1342\% \\
Polymeric nucleic acid & 1,472,613,528 & 0.3039\% \\
Glycan & 50,586,134 & 0.01044\% \\
Nucleotide/cofactor & 9,619,500 & 0.00199\% \\
Other ligand & 121,915,485 & 0.02516\% \\
Lipid/detergent & 17,296,940 & 0.00357\% \\
\bottomrule
\end{tabular}
\end{table}

\section{Hetero Analysis}
\label{app:audit}

\subsection{Analysis subset and hetero statistics}

The preliminary analysis uses a random sample of approximately 10\% of
the available data, yielding 1,868 successfully processed
map--structure pairs.
Among them, 1,458 pairs (78.05\%) contain nonzero hetero-labeled voxels,
showing that hetero density is common across the analyzed maps despite
its sparse voxel occupancy.
Of these pairs, 1,362 have successful atom and amino acid predictions
and are used for the candidate-neighborhood analysis.
The candidate-removal experiments use the class-specific successful
pairs defined below.

Table~\ref{tab:sparsity-detail} summarizes full-map voxel occupancy and
the proximity of non-protein residues to reference proteins.
Voxel fractions are pooled over all processed map voxels.
Among protein- and hetero-associated voxels, hetero voxels account for
14.06\%.

\begin{table}[htbp]
\caption{Full-map statistics for the 1,868-pair preliminary-analysis subset.
Voxel fractions are pooled over complete map grids.
The final three rows summarize residue-level proximity for 176,963
non-water, non-protein residues retained by the historical parser and
are not five-class prevalence estimates.}
\label{tab:sparsity-detail}
\centering
\small
\begin{tabular}{lr}
\toprule
Quantity & Value \\
\midrule
Analyzed map--structure pairs & 1,868 \\
Pairs with nonzero hetero-labeled voxels & 1,458 (78.05\%) \\
Protein-associated voxels / full grid & 0.709909\% \\
Hetero-associated voxels / full grid & 0.116131\% \\
Background voxels / full grid & 99.173960\% \\
Hetero / protein-plus-hetero voxels & 14.06\% \\
Residues with $d_{\mathrm{CA}}\leq4$~\angstrom{} & 17.1\% \\
Residues with $d_{\mathrm{CA}}>4$~\angstrom{},
$d_{\mathrm{heavy}}\leq3$~\angstrom{} & 14.1\% \\
Residues with $d_{\mathrm{CA}}>4$~\angstrom{},
$d_{\mathrm{heavy}}>3$~\angstrom{} & 68.8\% \\
\bottomrule
\end{tabular}
\end{table}

The final three rows use a separate residue-level denominator of 176,963
non-water, non-protein residues retained by the historical parser,
including ions and unknown residues.
For each residue, $d_{\mathrm{CA}}$ and $d_{\mathrm{heavy}}$ denote the
minimum distances from any retained non-hydrogen atom to a reference
protein C$\alpha$ atom and protein heavy atom, respectively.
The three groups are mutually exclusive: residues close to a reference
C$\alpha$, residues outside this range but close to another protein heavy
atom, and the remaining residues.
Protein is restricted to the 20 standard amino acids in ATOM records,
and HOH, WAT, and DOD are excluded as water.
Together, these statistics show that hetero-labeled voxels are sparse
across the full map but occur in most analyzed pairs, while a substantial
fraction of non-protein residues lie close to reference protein structure.

\subsection{Protein candidates near hetero components}

\paragraph{Analysis groups and residue centers.}
For the preliminary analysis, each hetero residue is represented by the
mean position of its retained non-hydrogen atoms.
Polymeric nucleic acids and glycans are represented residue by residue
rather than by a single center for the entire chain.
These residue centers are used only for the neighborhood analysis and
differ from the reference-atom anchors used to supervise \method.

The analysis uses fixed residue-name groups.
Nucleic acid includes A, C, G, U, DA, DC, DG, DT, and I;
glycan includes NAG, MAN, BMA, FUC, GAL, GLC, SIA, NDG, BGC, and XYS;
nucleotide/cofactor includes ATP, ADP, AMP, GTP, GDP, GNP, HEM, HEA,
FAD, FMN, NAD, NAP, SAH, SAM, COA, and FES; and lipid/detergent
includes LMT, POV, MC3, CLR, CHL, 3PE, PEE, PCF, BCR, CLA, DOP,
and CDL.
Other ligand contains the remaining residues after ions and unknown
components (UNK and UNL) are excluded.
This grouping is used only for the preliminary analysis.
Unlike the CCD- and entity-based taxonomy used for training, it does not
check polymer membership and may assign unlisted modified components to
the other-ligand group.

\paragraph{Protein candidates and matching.}
We form candidate protein sites from points with C$\alpha$ probability
$\geq0.5$.
Points are processed in descending probability order and greedily clustered
within 2~\angstrom{} of each seed; each point is assigned once, and the
probability-weighted cluster center defines one candidate.
Candidates are then matched to reference protein sites across the full map
without structural alignment.
Each candidate is associated with its nearest reference C$\alpha$, and
candidate--reference pairs are considered in ascending distance order.
A match is accepted when the distance is within 3~\angstrom{} and the
reference site has not already been used.
Accepted candidates are true positives (TP), and the remaining candidates
are false positives (FP); rejected candidates are not rematched to a
second-nearest reference site.
Reference protein sites are taken from C$\alpha$ atoms of the 20 standard
amino acids in ATOM records.

\paragraph{Neighborhood patterns.}
For each hetero residue center, we examine the surrounding
5~\angstrom{} neighborhood and record FP presence and protein support
independently.
Protein support is present when the neighborhood contains either a TP
candidate or a reference protein site, so a reference site missed by the
predictor still provides support.
These two indicators define four patterns: FP only, FP with protein
support, protein support only, and neither
(Table~\ref{tab:candidate-audit-retained}).

Neighborhoods can overlap.
Each hetero residue contributes one neighborhood, and the same protein
candidate may therefore occur in several neighborhoods.
The FP and TP occurrence counts in Table~\ref{tab:candidate-audit-retained}
include these repeated neighborhood occurrences rather than unique
candidates.
A map--structure pair may likewise contribute to more than one hetero
group.

\begin{table}[htbp]
\caption{Protein candidate patterns near hetero components in the
1,362 audited cases.
Percentages are pooled over residue neighborhoods within each analysis
group.
Protein support includes either a TP candidate or a reference protein
site.
FP and TP occurrences allow the same candidate to appear in overlapping
neighborhoods.
Rounding may cause small departures from 100\%.}
\label{tab:candidate-audit-retained}
\centering
\scriptsize
\resizebox{\linewidth}{!}{%
\begin{tabular}{lrrrrrrrr}
\toprule
Class & Pairs & Residues &
\shortstack{FP\\only} &
\shortstack{FP +\\protein\\support} &
\shortstack{Protein\\support\\only} &
Neither &
\shortstack{FP\\occurrences} &
\shortstack{TP\\occurrences} \\
\midrule
Nucleic acid & 264 & 105,890 & 78.4\% & 1.5\% & 0.1\% & 19.9\% & 501,865 & 1,326 \\
Glycan & 396 & 14,330 & 72.9\% & 6.6\% & 0.6\% & 20.0\% & 68,320 & 910 \\
Nucleotide/cofactor & 273 & 1,172 & 21.2\% & 71.2\% & 6.9\% & 0.6\% & 6,881 & 1,143 \\
Other ligand & 751 & 7,114 & 50.2\% & 44.0\% & 2.6\% & 3.2\% & 42,118 & 4,021 \\
Lipid/detergent & 207 & 6,007 & 77.3\% & 20.0\% & 0.4\% & 2.3\% & 40,676 & 1,246 \\
\bottomrule
\end{tabular}}
\end{table}

The resulting patterns differ substantially across hetero groups.
FP-only neighborhoods dominate around nucleic acids, glycans, and
lipids or detergents, whereas nucleotide/cofactor neighborhoods most
often contain both FP and protein support.
Other ligands show a more balanced mixture of these patterns.

\subsection{Candidate filtering and chain construction}

For each hetero group, we filter Cryo2Struct candidate states within
5~\angstrom{} of the annotated residue centers before chain construction.
These states follow the original HMM pipeline and are independent of the
candidate clustering used in the neighborhood analysis above.
The density map, atom and amino acid predictions, input sequences, and
decoding settings remain fixed.
After filtering, we rebuild the transition matrix using the retained states
and their corresponding emission rows, then evaluate the reconstructed
chains with the same matching protocol.
This comparison isolates the effect of filtering protein candidates near
hetero components on chain construction.

A class--map pair is included when the corresponding hetero group is
present and both the original and filtered runs succeed.
Runs contain at most 12,000 initial states and must retain enough states
for the sequence observations both before and after filtering.
Pairs are included even when no candidate state is removed, and successful
processing of the other hetero groups is not required.
Each group therefore has its own paired subset, and the subsets can overlap
across maps.

\begin{table}[htbp]
\caption{Effects of candidate filtering on chain construction for the
hetero groups defined above.
Win/loss reports the numbers of paired cases with increased or decreased
C$\alpha$ F$_1$ at 3~\AA{}; paired $n$ includes unchanged cases.
Net $\Delta$TP is the total change in matched protein sites, and removed
states/map is the mean number of filtered candidate states in each paired
subset.}
\label{tab:candidate-removal-retained}
\centering
\small
\begin{tabular}{lrrrr}
\toprule
Class & Paired $n$ & Win/loss & Net $\Delta$TP &
\shortstack{Removed\\states/map} \\
\midrule
Nucleic acid & 102 & 84/17 & +4,620 & 961 \\
Glycan & 172 & 96/70 & +687 & 104 \\
Nucleotide/cofactor & 84 & 40/41 & +200 & 34 \\
Other ligand & 357 & 185/158 & +389 & 45 \\
Lipid/detergent & 111 & 65/44 & +523 & 78 \\
\bottomrule
\end{tabular}
\end{table}

Candidate filtering produces both improvements and degradations in every
hetero group (Table~\ref{tab:candidate-removal-retained}).
The effect is most pronounced for nucleic acids, where most paired cases
improve and the total number of matched protein sites increases
substantially, but the remaining groups also show mixed outcomes.
These results indicate that filtering can remove misleading protein
candidates, while uniform filtering can also discard candidates that are
useful for chain construction.

\section{Architecture Details}
\label{app:architecture}
\label{app:code_architecture}

\subsection{Hetero detector}
\label{app:hetero_detector}

The hetero detector consists of a local 3D SCUNet, a global Transformer
at the coarsest scale, and an eight-class prediction head.
Table~\ref{tab:composition_architecture} summarizes the feature dimensions
for a $48^3$ density crop.

\begin{table}[htbp]
\caption{Architecture of the hetero detector.
Spatial scales are downsampling factors relative to the input crop.}
\label{tab:composition_architecture}
\centering
\small
\begin{tabular}{lll}
\toprule
Stage & Feature shape & Operation \\
\midrule
Input & $1\times48^3$ & Normalized density \\
Scale 1 & $32\times48^3$ & Stem and local blocks \\
Scale 2 & $64\times24^3$ & Downsampling and local blocks \\
Scale 4 & $128\times12^3$ & Downsampling and local blocks \\
Scale 8 & $256\times6^3$ & Bottleneck \\
Global context & $384\times6^3$ & 10 Transformer blocks \\
Decoder & Scales $8\rightarrow4\rightarrow2\rightarrow1$
        & Upsampling with modulated skips \\
Refinement & $64\times48^3$ & Conv3d, GroupNorm, SiLU \\
Prediction & $8\times48^3$ & Voxel classification head \\
\bottomrule
\end{tabular}
\end{table}

Algorithm~\ref{alg:hetero_detector} summarizes the detector forward pass.
The local backbone first produces features at four spatial scales.
Global context is introduced at the coarsest scale and used to modulate
the skip features before decoding.
The resulting feature pyramid is also exposed to the backbone predictor.

\begin{algorithm}[htbp]
\caption{Forward pass of the hetero detector}
\label{alg:hetero_detector}
\small
\begin{algorithmic}[1]
\Require Normalized density crop $X$; detector parameters $\theta_h$
\Ensure Class probabilities $p$ and hetero features
$\{F_s^h\}_{s\in\{8,4,2,1\}}$

\State $S_1 \gets \operatorname{Stem}_{\theta_h}(X)$
\State $S_2 \gets \operatorname{Down}_2(S_1)$;
       $S_4 \gets \operatorname{Down}_4(S_2)$;
       $S_8 \gets \operatorname{Down}_8(S_4)$
\State $B \gets \operatorname{Body}(S_8)$
\State $(B^g,S_4^g,S_2^g,S_1^g)
       \gets \operatorname{GlobalContext}(B,S_4,S_2,S_1)$
\State $F_8^h \gets B^g + S_8$
\State $F_4^h \gets \operatorname{Decode}_4(F_8^h) + S_4^g$
\State $F_2^h \gets \operatorname{Decode}_2(F_4^h) + S_2^g$
\State $H_1 \gets \operatorname{Decode}_1(F_2^h) + S_1^g$
\State $F_1^h \gets \operatorname{Refine}(H_1)$
\State $p \gets
       \operatorname{softmax}(\operatorname{Predict}(F_1^h))$
\State \Return $p,\{F_8^h,F_4^h,F_2^h,F_1^h\}$
\end{algorithmic}
\end{algorithm}

The local backbone is a 3D adaptation of the SCUNet design
\citep{zhang2023practical,liu2021swin}.
It contains seven stages, with two Conv--Transformer blocks at each stage.
Each block first projects and splits its channels into convolutional and
Transformer branches.
The convolutional branch applies two $3^3$ convolutions with filter
response normalization and a residual connection.
The Transformer branch applies window self-attention followed by an MLP;
both operations use pre-normalization and residual updates.
The two branches are then concatenated, projected, and added back to the
block input.
Window attention uses $3^3$ windows with head dimension 16, with regular
and shifted windows alternating between successive blocks.

After the local bottleneck is computed, a global Transformer processes all
spatial positions at the coarsest scale.
The 256-channel bottleneck is projected to 384 channels, giving 216 tokens
on the $6^3$ grid.
Ten Transformer blocks process these tokens with eight attention heads,
three-dimensional rotary positional embeddings, and SwiGLU feed-forward
layers.
The resulting global context is added to the bottleneck and also modulates
the skip features used during decoding.

Let $B$ denote the local bottleneck, $C$ the global context, and $S_s$
a skip feature at scale $s\in\{4,2,1\}$.
The global context is incorporated as
\begin{equation}
\begin{aligned}
C&=\mathcal G_h(W_{\mathrm{in}}B),
&
B'&=B+W_oC,\\
S_s'
&=
S_s\odot
\left[
1+\tanh\!\left(
G_s(\operatorname{Up}_s C)
\right)
\right],
&&s\in\{4,2,1\}.
\end{aligned}
\label{eq:composition_global_context}
\end{equation}
Here, $W_{\mathrm{in}}$ and $W_o$ project the bottleneck into and out of
the global Transformer.
$\operatorname{Up}_s$ resizes the global context to the corresponding
skip resolution, and $G_s$ maps it to the skip feature dimension.
The output projection and skip gates are zero initialized, so the global
branch initially leaves the local backbone unchanged.

After decoding, a refinement layer maps the full-resolution feature from
32 to 64 channels.
The prediction head outputs eight logits for background, protein backbone,
protein side chain, polymeric nucleic acid, glycan, nucleotide/cofactor,
other ligand, and lipid/detergent.

For backbone localization, we expose four detector features, denoted
$F_s^h$.
$F_8^h$ is the context-updated coarse feature, $F_4^h$ and $F_2^h$ are
decoder features at the corresponding scales, and $F_1^h$ is the refined
full-resolution feature.

The detector is trained with focal cross-entropy and foreground Dice loss.
Let $\mathcal V$ denote the non-ignored voxels, $y_v$ the target class at
voxel $v$, and $p_{vc}$ the predicted probability for class $c$.
The objective is
\begin{equation}
\mathcal L_h
=
-\frac{1}{|\mathcal V|}
\sum_{v\in\mathcal V}
(1-p_{v,y_v})^2
\log p_{v,y_v}
+
0.5\,\mathcal L_{\mathrm{Dice}},
\label{eq:detector_loss}
\end{equation}
where
\begin{equation}
\mathcal L_{\mathrm{Dice}}
=
1-
\frac{1}{|\mathcal C_+|}
\sum_{c\in\mathcal C_+}
\frac{
2\sum_{v\in\mathcal V}
p_{vc}\mathbf 1[y_v=c]+\epsilon
}{
\sum_{v\in\mathcal V}p_{vc}
+
\sum_{v\in\mathcal V}\mathbf 1[y_v=c]
+\epsilon
}.
\label{eq:detector_dice_appendix}
\end{equation}
Here, $\mathcal C_+$ contains the foreground classes present in the
distributed batch, and $\epsilon=10^{-6}$.
The background class is excluded from the class average, while all
non-ignored voxels, including background voxels, contribute to each
foreground Dice denominator.
Ignored voxels are excluded from both loss terms.
When no foreground class is present, the Dice term is set to zero.


\subsection{Backbone predictor and multiscale fusion}
\label{app:backbone_architecture}
\label{app:joint_fusion}

The backbone predictor follows the same local 3D SCUNet architecture as
the hetero detector, but uses independent parameters.
It predicts three backbone response maps and receives hetero features at
scales $s\in\{8,4,2,1\}$.
At each scale, a hetero update $\Delta_s$ is added to the protein feature,
\begin{equation}
\widetilde F_s^p=F_s^p+\Delta_s.
\label{eq:backbone_feature_flow_appendix}
\end{equation}
Table~\ref{tab:fusion_sites} gives the feature dimensions and fusion
operation at each scale.

\begin{table}[htbp]
\caption{Multiscale hetero fusion in the backbone predictor.
Shapes give channels and spatial dimensions for a $48^3$ input crop.}
\label{tab:fusion_sites}
\centering
\small
\begin{tabular}{llll}
\toprule
Scale $s$ & Protein $F_s^p$ & Hetero $F_s^h$ & Fusion \\
\midrule
8 & $256\times6^3$ & $256\times6^3$ & Global cross-attention \\
4 & $128\times12^3$ & $128\times12^3$ & Window cross-attention \\
2 & $64\times24^3$ & $64\times24^3$ & Gated convolution \\
1 & $32\times48^3$ & $64\times48^3$ & Gated convolution \\
\bottomrule
\end{tabular}
\end{table}

\paragraph{Coarse-scale cross-attention.}
At scales 8 and 4, protein features provide queries and hetero features
provide keys and values.
With $m_s$ denoting the spatial validity mask,
\begin{equation}
\Delta_s
=
m_s\odot
W_s^O
\operatorname{Attn}_s\!\left(
Q_s(F_s^p),
K_s(F_s^h),
V_s(F_s^h)
\right),
\qquad s\in\{8,4\}.
\label{eq:backbone_cross_attention}
\end{equation}
Each projection applies layer normalization followed by a learned linear
mapping.
At scale 8, eight heads of dimension 32 attend over all 216 spatial
tokens.
At scale 4, eight heads of dimension 16 attend within non-overlapping
$3^3$ windows.
Invalid keys are masked during attention, invalid query updates are
zeroed, and coarse validity masks are obtained by max pooling the input
mask.

\paragraph{Fine-scale gated fusion.}
At scales 2 and 1, the hetero feature is first projected to the protein
feature dimension.
A spatial gate predicted from both streams then controls its contribution:
\begin{align}
U_s
&=
\operatorname{SiLU}\!\left[
\operatorname{GN}_8\!\left(
\operatorname{Conv}_{3\times3\times3}(F_s^h)
\right)
\right],
\nonumber\\
G_s
&=
\sigma\!\left(
\operatorname{Conv}_{1\times1\times1}
([F_s^p;U_s])
\right),
\nonumber\\
\Delta_s
&=
m_s\odot
\operatorname{Conv}^{\mathrm{out}}_{3\times3\times3}
(G_s\odot U_s),
\qquad s\in\{2,1\}.
\label{eq:backbone_gated_fusion}
\end{align}
The projection preserves 64 channels at scale 2 and maps the
64-channel hetero feature to 32 channels at scale 1.
Attention output projections and the final residual convolutions are
zero initialized, so the hetero fusion initially contributes no residual
update.

The hetero detector remains fixed during backbone training, so gradients
from the backbone predictor do not update the detector.
The final fused protein feature is mapped to three continuous response
maps ordered as N, C$\alpha$, and C.

\paragraph{Training objective.}
For response channel $a$, $\mathcal V_a^+$ contains valid central-region
voxels with target response $A_a(v)>2$ on the $[0,100]$ scale, and
$\mathcal V_a^-$ contains the remaining valid voxels.
Equation~\ref{eq:backbone_loss_main} gives equal weight to these two groups
and then averages equally over the three response channels.
An empty group contributes zero; errors and valid counts are aggregated
across devices before normalization.

\paragraph{Full-map backbone localization.}
We evaluate overlapping $48^3$ crops at stride 16 and average their valid
backbone responses at each map voxel.
Responses are set to zero where the normalized input density is at or
below $10^{-6}$.
To extract initial protein sites, we threshold the averaged C$\alpha$
response at 20 on the $[0,100]$ scale, followed by density-weighted
mean-shift and peak merging.
The mean-shift resolution parameter is 6~\angstrom{}, the maximum shift
distance is 10~\angstrom{}, and peaks within 1.5~\angstrom{} are merged.

\paragraph{Amino acid type prediction.}
\label{app:amino_acid_prediction}

The amino acid predictor is an independent 3D SCUNet with a 20-class
output head.
It maps $48^3$ density crops to voxel-wise residue-type probabilities.
Training uses cross-entropy over voxels with valid amino acid labels:
\begin{equation}
\mathcal L_{\mathrm{AA}}
=
-\frac{1}{|\mathcal V_{\mathrm{AA}}|}
\sum_{v\in\mathcal V_{\mathrm{AA}}}
\log p^a_{v,y_v^a},
\label{eq:aa_loss}
\end{equation}
where $\mathcal V_{\mathrm{AA}}$ contains valid protein voxels,
$y_v^a$ is the reference residue type, and $p^a_{v,y_v^a}$ is its
predicted probability.
Residue labels are assigned from the nearest protein heavy atom within
2~\angstrom{}; voxels without a valid residue label are ignored.

During structure refinement, the predicted residue-type logits are sampled
at the refined protein sites to form fragment profiles.
Alignment of the input protein sequences to these profiles assigns residue
identities and sequence positions for final assembly.

\subsection{Structure refinement with hetero candidates}
\label{app:protein_architecture}
\label{app:hetero_candidate_extraction}

The structure stage takes the density map, initial protein sites, and
hetero predictions from the detector.
Hetero predictions are first converted into sparse candidates and encoded
as fixed context for structure refinement.

\paragraph{Candidate extraction and encoding.}
Let $z_c(v)$ denote the detector logit for class $c$, and let
$\mathcal C_h$ contain the five hetero classes.
We compute the total hetero confidence as
\begin{equation}
\ell(v)
=
\operatorname{LSE}_{c\in\mathcal C_h}z_c(v)
-
\operatorname{LSE}_{c\notin\mathcal C_h}z_c(v),
\qquad
o(v)=\sigma(\ell(v)),
\label{eq:hetero_candidate_logodds}
\end{equation}
where $\operatorname{LSE}$ denotes log-sum-exp.
Candidates are $3^3$ local maxima of $\ell(v)$.
To avoid duplicates from overlapping crops, each voxel contributes
candidates only through its assigned central crop region.

Candidate $j$ stores its position $y_j$, confidence $o_j=o(y_j)$, and
conditional class probabilities
\begin{equation}
\pi_j
=
\operatorname{softmax}\!\left(z_{\mathcal C_h}(y_j)\right).
\end{equation}
Its input descriptor is
\begin{equation}
\begin{aligned}
x_j
&=
[\bar\ell_j;\,o_j;\,\pi_j;\,E_j;\,\max_c\pi_{jc}],\\
\bar\ell_j
&=
\operatorname{clip}(\ell_j,-12,12)/12,\\
E_j
&=
-\frac{1}{\log 5}
\sum_c
\pi_{jc}\log\max(\pi_{jc},10^{-8}),
\end{aligned}
\label{eq:hetero_token_input}
\end{equation}
which combines class information, confidence, and prediction uncertainty.
A shared encoder maps $x_j$ to a 128-dimensional embedding $e_j$.

For each initial protein site $i$, we select up to 32 candidates within
12~\angstrom{}, ranked by confidence, to form $\mathcal N_i$.
Candidate positions and neighborhood membership remain fixed throughout
refinement.

\paragraph{Protein features and frames.}
The density initializer extracts local density around each protein site and
between neighboring sites.
Convolutional encoders produce 256-dimensional residue features $H^0$ and
128-dimensional pair features $Z^0$.
Each protein site also carries a rigid frame $T_i=(R_i,t_i)$.

Sixteen three-track structure blocks jointly update residue features,
pair features, and frames.
The blocks combine residue attention, pair updates, and invariant point
attention (IPA) \citep{jumper2021highly}.
Hetero conditioning is applied before blocks 4, 8, 12, and 16:
\begin{equation}
(H^\ell,Z^\ell,T^\ell)
=
\mathcal S_\ell(
\widetilde H^{\ell-1},
Z^{\ell-1},
T^{\ell-1}),
\qquad
\ell=1,\ldots,16,
\label{eq:protein_refinement_flow}
\end{equation}
where $\widetilde H=H$ at blocks without hetero conditioning.

\paragraph{Frame-relative hetero conditioning.}
Although hetero candidates remain fixed in the map, their relation to a
protein site changes as its frame is updated.
For frame $T_i=(R_i,t_i)$ and candidate $y_j$, we compute
\begin{equation}
d_{ij}=\|y_j-t_i\|_2,
\qquad
u_{ij}
=
\frac{R_i^\top(y_j-t_i)}
{\max(d_{ij},10^{-6})}.
\label{eq:hetero_geometry_appendix}
\end{equation}
These quantities are recomputed from the current frames at every
conditioning step.

Distance is expanded with 16 Gaussian radial basis functions,
\begin{equation}
r_m(d)
=
\exp\!\left[
-\frac12
\left(
\frac{d-12m/15}{0.75}
\right)^2
\right],
\qquad
m=0,\ldots,15.
\label{eq:hetero_distance_rbf}
\end{equation}
The radial features, local direction, and log confidence are mapped to one
additive bias per attention head.

Protein features provide queries, while candidate embeddings provide keys
and values.
For one attention head,
\begin{equation}
\begin{aligned}
\alpha_{ij}
&=
\operatorname{softmax}_{j\in\mathcal N_i}
\left(
\frac{q_i^\top k_j}{\sqrt{d_h}}
+
b(d_{ij},u_{ij},o_j)
\right),\\
c_i
&=
\sum_{j\in\mathcal N_i}
\alpha_{ij}v_j .
\end{aligned}
\label{eq:protein_hetero_attention}
\end{equation}
Each conditioning module uses four heads of dimension 32.
Invalid candidates are masked before attention normalization.

The head outputs are concatenated into $c_i^{\mathrm{all}}$.
A gate then controls how much hetero context is added to each protein site:
\begin{equation}
\begin{aligned}
n_i &= |\mathcal N_i|/32,
\qquad
o_i^{\max} = \max_{j\in\mathcal N_i}o_j,\\
\gamma_i
&=
\sigma\!\left(
g([
\operatorname{LN}(h_i);
c_i^{\mathrm{all}};
n_i;
o_i^{\max}])
\right),\\
\widetilde h_i
&=
h_i+\gamma_iP_O(c_i^{\mathrm{all}}).
\end{aligned}
\label{eq:hetero_gate_appendix}
\end{equation}
The gate uses the protein feature, hetero context, neighborhood size, and
maximum candidate confidence.
$P_O$ is zero initialized, so hetero conditioning starts from a zero
residual.
Sites without valid candidates receive no hetero update.

\paragraph{Torsions and sequence assignment.}
The final residue features predict torsions, while the refined frames define
the backbone geometry.
These predictions first form provisional chain fragments.
Residue-type logits sampled at the refined protein sites provide fragment
profiles for alignment to the input sequences.
The resulting sequence assignments provide residue identities and positions
for fragment assembly and final all-atom construction.

\paragraph{Training objective.}
The current structure model is trained with backbone-coordinate and torsion
supervision only.
For valid backbone atoms,
\begin{equation}
\mathcal L_{\mathrm{backbone}}
=
\operatorname*{mean}_{(i,a)\in\mathcal M_{\mathrm{bb}}}
\rho\!\left(
\|\widehat x_{ia}-x_{ia}\|_2
\right),
\label{eq:structure_backbone_loss}
\end{equation}
where $\rho$ is the radial Huber loss,
\begin{equation}
\rho(d)=
\begin{cases}
d^2/2, & d<1,\\
d-1/2, & d\geq1.
\end{cases}
\label{eq:structure_huber_loss}
\end{equation}
For valid torsions,
\begin{equation}
\mathcal L_{\mathrm{torsion}}
=
\operatorname*{mean}_{(i,k)\in\mathcal M_{\mathrm{tor}}}
\min\!\left(
\|\widehat z_{ik}-z_{ik}\|_2^2,
\|\widehat z_{ik}-z_{ik}^{\mathrm{alt}}\|_2^2
\right),
\label{eq:structure_torsion_loss}
\end{equation}
where $\widehat z_{ik}$ is the normalized predicted sine--cosine pair and
$z_{ik}^{\mathrm{alt}}$ represents an equivalent torsion target when one
exists.
The final objective is
\begin{equation}
\mathcal L_{\mathrm{structure}}
=
\mathcal L_{\mathrm{backbone}}
+
0.1\,\mathcal L_{\mathrm{torsion}}.
\label{eq:structure_loss}
\end{equation}

\subsection{End-to-end inference}
\label{app:inference_algorithm}

Algorithm~\ref{alg:cryocue_inference} summarizes the complete inference
pipeline.
The hetero detector $D$, backbone predictor $B$, and amino acid predictor
$A$ first process the density map.
The resulting protein sites and hetero candidates are passed to the
structure network for three rounds of refinement.
Each round starts from the protein frames produced by the previous round,
while the hetero candidates remain fixed in the map.
Their geometry relative to the current protein frames is recomputed during
each round.

\begin{algorithm}[htbp]
\caption{Inference with \method}
\label{alg:cryocue_inference}
\small
\begin{algorithmic}[1]
\Require Normalized map $M$, protein sequences $Q$; trained models
$D,B,A,\mathcal S$
\Ensure Reconstructed protein structure $\widehat S$

\For{$X\in\operatorname{Crops}_{48,16}(M)$}
    \State $(p_X,F_X^h)\gets D(X)$
    \State $R_X\gets B(100X,F_X^h)$
    \State $X_a\gets\operatorname{AAInput}(X)$;
    $\Lambda_X\gets A(X_a)$
    \State $\mathcal K_X\gets\operatorname{Candidates}(p_X)$
\EndFor

\State $C\gets
\operatorname{ExtractSites}(\operatorname{OverlapMean}(\{R_X\}))$
\State $\mathcal K\gets\bigcup_X\mathcal K_X$
\State $T^{(0)}\gets\operatorname{InitializeFrames}(C)$
\State $\{e_j\}\gets\operatorname{EncodeCandidates}(\mathcal K)$
\State $\mathcal N_i\gets
\operatorname{SelectNeighbors}(C_i,\mathcal K)$

\For{$r=1,\ldots,3$}
    \State $(T^{(r)},U^{(r)})\gets
    \Call{Refine}{M,T^{(r-1)},\mathcal K,\{e_j\},\{\mathcal N_i\}}$
\EndFor

\State $T\gets T^{(3)}$;
       $U\gets U^{(3)}$
\State $\mathcal F\gets\operatorname{Trace}(T)$
\State $\Pi\gets\operatorname{Profiles}(\{\Lambda_X\},T,\mathcal F)$
\State $\mathcal A\gets\operatorname{HMMAlign}(\Pi,Q)$
\State $(\mathcal J,a)\gets
\operatorname{AssemblePrune}(\mathcal F,\mathcal A,T)$
\State $\widehat S\gets
\operatorname{BuildAtoms}(T[\mathcal J],U[\mathcal J],a)$
\State \Return $\widehat S$

\Statex
\Function{Refine}{$M,T^0,\mathcal K,\{e_j\},\{\mathcal N_i\}$}
    \State $(H^0,Z^0)\gets\operatorname{InitializeFeatures}(M,T^0)$

    \For{$\ell=1,\ldots,16$}
        \State $\widetilde H\gets H^{\ell-1}$
        \If{$\ell\in\{4,8,12,16\}$}
            \State $\mathcal G\gets
            \operatorname{RelativeGeometry}(T^{\ell-1},
            \mathcal K,\{\mathcal N_i\})$
            \State $\widetilde H\gets
            \operatorname{HeteroCondition}_{\ell}
            (H^{\ell-1},\{e_j\},\mathcal G,\{\mathcal N_i\})$
        \EndIf
        \State $(H^\ell,Z^\ell,T^\ell)\gets
        \mathcal S_\ell(\widetilde H,Z^{\ell-1},T^{\ell-1})$
    \EndFor

    \State $U\gets\operatorname{TorsionHead}(H^{16})$
    \State \Return $T^{16},U$
\EndFunction
\end{algorithmic}
\end{algorithm}

$\operatorname{Crops}_{48,16}$ extracts $48^3$ crops at stride 16, and
$X_a$ denotes the density input prepared for the independently trained amino
acid predictor. Candidate extraction uses the central crop region so that
overlapping crops do not produce duplicate candidates.
$\mathcal F$ denotes provisional chain fragments, $\Pi$ their residue-type
profiles, and $a$ the residue identities assigned by sequence alignment.

\section{Training Details}
\label{app:training}
\label{app:joint_training}

\subsection{Training procedure}

Training proceeds in four stages.
We first train the hetero detector with anchor supervision.
The detector is then fixed while the backbone predictor and multiscale
fusion modules are trained for backbone localization.
The amino acid predictor is trained independently from density and does
not use hetero features.
Finally, we train the structure network with predicted hetero candidates
from the fixed detector.

Algorithm~\ref{alg:cryocue_training} summarizes the training procedure.
All learnable models are trained from scratch without pretrained weights.
Each model has its own parameters, optimizer, and training objective.

\begin{algorithm}[htbp]
\caption{Staged training of \method}
\label{alg:cryocue_training}
\small
\begin{algorithmic}[1]
\Require Training data $\mathcal D$; model parameters
$\theta_h,\theta_b,\theta_a,\theta_s$
\Ensure Trained parameters $\theta_h,\theta_b,\theta_a,\theta_s$

\Statex $\triangleright$ \textit{Hetero representation}
\For{$t=1,\ldots,N_h$}
    \State $(X,Y_h,m_h)\sim\mathcal D_h$
    \State $(p,F^h)\gets D_{\theta_h}(X)$
    \State $L\gets\mathcal L_h(p,Y_h;m_h)$
    \State $\theta_h\gets
    \operatorname{Step}_h(\theta_h,\nabla_{\theta_h}L)$
\EndFor
\State $D_{\theta_h}\gets\operatorname{FreezeEval}(D_{\theta_h})$

\Statex $\triangleright$ \textit{Backbone localization}
\For{$t=1,\ldots,N_b$}
    \State $(X,Y_b,m_b)\sim\mathcal D_b$
    \State $(p,F^h)\gets D_{\theta_h}(X)$
    \State $\widehat Y_b\gets B_{\theta_b}(100X,F^h)$
    \State $L\gets
    \mathcal L_{\mathrm{voxel}}(\widehat Y_b,Y_b;m_b)$
    \State $\theta_b\gets
    \operatorname{Step}_b(\theta_b,\nabla_{\theta_b}L)$
\EndFor
\State $B_{\theta_b}\gets\operatorname{FreezeEval}(B_{\theta_b})$

\Statex $\triangleright$ \textit{Amino acid prediction}
\For{$t=1,\ldots,N_a$}
    \State $(X_a,Y_a,m_a)\sim\mathcal D_a$
    \State $p^a\gets\operatorname{softmax}(A_{\theta_a}(X_a))$
    \State $L\gets\mathcal L_{\mathrm{AA}}(p^a,Y_a;m_a)$
    \State $\theta_a\gets
    \operatorname{Step}_a(\theta_a,\nabla_{\theta_a}L)$
\EndFor
\State $A_{\theta_a}\gets\operatorname{FreezeEval}(A_{\theta_a})$

\Statex $\triangleright$ \textit{Structure refinement}
\For{$e=1,\ldots,E_s$}
    \For{$(M,T^0,\mathcal K,Y_s)\sim\mathcal D_s^{(e)}$}
        \State $(T,U)\gets S_{\theta_s}(M,T^0,\mathcal K)$
        \State $L\gets\mathcal L_{\mathrm{structure}}(T,U;Y_s)$
        \State $\theta_s\gets
        \operatorname{Step}_s(\theta_s,\nabla_{\theta_s}L)$
    \EndFor
\EndFor

\State \Return $(\theta_h,\theta_b,\theta_a,\theta_s)$
\end{algorithmic}
\end{algorithm}

Here, $\mathcal D_h$, $\mathcal D_b$, and $\mathcal D_a$ use the
voxel-stage sampling described below.
For structure training, $\mathcal D_s^{(e)}$ contains the density map,
initial protein frames, predicted hetero candidates, and reference
structure targets for pass $e$.

\subsection{Voxel-stage training}

The hetero detector, backbone predictor, and amino acid predictor use
$48^3$ density crops.
Training requests follow a 94/4/2 mixture of natural crops,
hetero-positive crops, and protein-containing crops.
The natural stream provides broad map coverage, while the other two
streams increase exposure to hetero and protein regions.
The hetero-positive stream also maintains coverage across the five hetero
classes.

Density and supervision receive the same orientation-preserving cube
rotation, selected from the 24 proper cube rotations.
Reflections are not used.
Detector supervision covers all valid positions in the $48^3$ crop.
Backbone and amino acid supervision use the valid central $16^3$ region.

The hetero detector, backbone predictor, and amino acid predictor are
trained from scratch with independent parameters.
Detector training jointly optimizes the local backbone, global context
module, and prediction head.
After detector training, its parameters are fixed while the backbone
predictor and multiscale fusion modules are optimized for backbone
localization.

The detector receives normalized density $X$, while the backbone predictor
receives $100X$.
This scaling changes only the input amplitude and leaves the map
coordinates unchanged.

Table~\ref{tab:training-configuration} summarizes the optimization
settings for the three voxel models.

\begin{table}[htbp]
\caption{Optimization settings for the voxel-stage models.
Learning rates are peak values and training duration is measured in
optimizer updates.}
\label{tab:training-configuration}
\centering
\small
\begin{tabular}{llll}
\toprule
Setting & Hetero detector & Backbone predictor & Amino acid predictor \\
\midrule
GPUs 
    & 8
    & 32
    & 32 \\
Per-device batch
    & 32
    & 16
    & 16 \\
Global batch size
    & 256
    & 512
    & 512 \\
Optimizer updates
    & 225,170
    & 95,231
    & 95,231 \\
Learning rate
    & $10^{-4}$
    & $10^{-4}$
    & $10^{-4}$ \\
Warmup updates
    & 1,000
    & 500
    & 500 \\
\bottomrule
\end{tabular}
\end{table}

The voxel models use AdamW~\citep{loshchilov2017decoupled} with $\beta=(0.9,0.999)$, $\epsilon=10^{-8}$, and gradient clipping at norm 1.0.
Learning rates use linear warmup followed by cosine decay.
Weight decay is 0.01 for detector parameters and for the backbone trunk
and fusion modules. Bias, normalization, and other one-dimensional
parameters use no weight decay where applicable.
The amino acid predictor uses the same learning-rate schedule as the
backbone predictor but has no hetero-fusion parameter group.

The amino acid predictor has independent parameters and its own optimizer.
It is trained from density only, using cross-entropy over valid residue
type labels.
Its predictions are not used for backbone localization or structure
refinement; they are used later for sequence assignment and chain
construction.

\subsection{Structure-stage training}

Structure training uses perturbed reference backbone geometry to initialize
the protein frames, providing a controlled denoising task for learning
structure refinement.
For each training case and epoch, we uniformly sample one reference residue
as the center and take its 128 spatially nearest residues.
Independent Gaussian noise with standard deviation 1~\angstrom{} is added
to their reference backbone coordinates before the initial frames are
constructed.
With probability 0.5, the frame rotations are replaced by random rotations
while their translations are retained.
The unperturbed reference structure provides the coordinate and torsion
targets.
This initialization is used only during training.
At inference, structure refinement instead starts from the protein sites
predicted by backbone localization.

Hetero information is predicted rather than taken from reference hetero
annotations.
The fixed detector provides the hetero candidates and their descriptors.
For each initial protein frame, up to 32 candidates within
12~\angstrom{} are selected as described in
Appendix~\ref{app:protein_architecture}.
Candidate positions and neighborhood membership remain fixed during each
forward pass, while their geometry is recomputed as the protein frames
change.

The structure network is trained from scratch.
The shared candidate encoder and four hetero-conditioning modules are
integrated with the three-track structure backbone described in
Section~\ref{sec:protein_reconstruction}.
The structure backbone, torsion head, candidate encoder, and conditioning
modules are optimized jointly.
Training uses a single refinement forward pass without recycling.

The structure-training inventory contains 17,820 records.
Training uses eight GPUs with one example per device and no gradient
accumulation, giving a global batch size of 8.
Network computation uses BF16, while frame composition and geometric loss
computation use FP32.

We train for 24 epochs.
Each epoch contains 2,228 optimizer updates, giving 53,472 updates in total.
We use AdamW with $\beta=(0.9,0.999)$, $\epsilon=10^{-8}$, and gradient
clipping at norm 1.0.
The peak learning rate is $10^{-4}$ for all trainable parameters.
Matrix parameters use weight decay $10^{-4}$; bias, normalization, and
other one-dimensional parameters use no weight decay.
The learning rate uses 200 warmup updates followed by cosine decay with a
minimum learning-rate ratio of 0.1.

The structure objective contains only the backbone-coordinate and torsion
losses defined in Section~\ref{sec:protein_reconstruction}.


\section{Evaluation Details}
\label{app:evaluation_metrics}

For each map, let
$P=\{\mathbf p_i\}_{i=1}^{N_P}$ and
$R=\{\mathbf r_j\}_{j=1}^{N_R}$
denote the predicted and reference protein C$\alpha$ coordinates,
respectively.
Only protein residues recognized by the fixed scoring implementation
are included.

\paragraph{TM-score.}
We use protein-only US-align to compare the reconstructed and reference
structures \citep{zhang2022us}.
The reported TM-score is normalized by the reference protein length.
US-align returns a residue correspondence $\mathcal A$ and an optimized
rigid transformation.
The score is
\begin{equation}
\mathrm{TM\text{-}score}
=
\frac{1}{N_R}
\sum_{(i,j)\in\mathcal A}
\frac{1}{
1+\left(d^{\mathrm{ali}}_{ij}/d_0(N_R)\right)^2
},
\label{eq:metric_tm}
\end{equation}
where $d^{\mathrm{ali}}_{ij}$ is the C$\alpha$ distance after structural
alignment and $d_0(N_R)$ is the default reference-length-dependent
distance scale used by US-align.
We use the backend score reported with the predicted structure as
Structure~1 and the reference structure as Structure~2 in complex
alignment mode (\texttt{-mm 1}).

\paragraph{Local Distance Difference Test (lDDT).}
We use a C$\alpha$-only implementation of the local Distance Difference
Test (lDDT) \citep{mariani2013lddt}.
Let
\begin{equation}
\mathcal D
=
\left\{
(j,k):
j<k,\;
\|\mathbf r_j-\mathbf r_k\|_2<15~\text{\AA}
\right\}
\end{equation}
contain all reference C$\alpha$ pairs within 15~\angstrom{}.
For a reference residue $j$, let $a(j)$ denote its corresponding predicted
residue under the US-align correspondence.
When both endpoints are mapped, their distance error is
\begin{equation}
\Delta_{jk}
=
\left|
\|\mathbf p_{a(j)}-\mathbf p_{a(k)}\|_2
-
\|\mathbf r_j-\mathbf r_k\|_2
\right|.
\end{equation}
We compute
\begin{equation}
\mathrm{lDDT}
=
\frac{1}{4|\mathcal D|}
\sum_{(j,k)\in\mathcal D}
\sum_{t\in\{0.5,1,2,4\}}
\mathbf 1[a(j),a(k)\text{ exist}]
\mathbf 1[\Delta_{jk}<t].
\label{eq:metric_lddt}
\end{equation}
Pairs with an unmapped endpoint contribute zero.
Both intra-chain and inter-chain reference pairs are included.
No additional coordinate superposition is applied after the US-align
correspondence is established.

\paragraph{Coverage.}
Coverage measures the fraction of reference residues included in the
US-align structural correspondence:
\begin{equation}
\mathrm{Coverage}
=
100\,
\frac{|\mathcal A|}{N_R}.
\label{eq:metric_coverage}
\end{equation}
Coverage is determined by structural alignment and does not impose the
3~\angstrom{} positional criterion used for backbone localization below.
It therefore differs from backbone recall.

\paragraph{Backbone precision and recall.}
Backbone localization is evaluated directly in the original cryo-EM map
coordinate frame using C$\alpha$ coordinates.
We enumerate all predicted--reference pairs within 3~\angstrom{}, sort
them by increasing distance, and greedily accept a pair when neither
endpoint has been matched previously.
This produces a one-to-one matching $\mathcal M_3$ without structural
superposition.
Precision and recall are
\begin{equation}
\mathrm{Precision}
=
100\,
\frac{|\mathcal M_3|}{N_P},
\qquad
\mathrm{Recall}
=
100\,
\frac{|\mathcal M_3|}{N_R}.
\label{eq:metric_backbone_pr}
\end{equation}
The precision reported here is the position-based quantity previously
referred to as backbone accuracy, while recall corresponds to backbone
completeness.
The matching is distance-ordered greedy matching rather than a
maximum-cardinality or Hungarian assignment.

\paragraph{Backbone RMSD.}
RMSD measures coordinate error among the backbone sites accepted by
$\mathcal M_3$:
\begin{equation}
\mathrm{RMSD}
=
\left(
\frac{1}{|\mathcal M_3|}
\sum_{(i,j)\in\mathcal M_3}
\|\mathbf p_i-\mathbf r_j\|_2^2
\right)^{1/2}.
\label{eq:metric_backbone_rmsd}
\end{equation}
No additional rigid-body fitting is performed.
Unmatched sites are excluded, so RMSD measures localization error
conditional on successful recovery rather than reconstruction completeness.

\paragraph{Backbone break rate.}
We construct reference and predicted backbone edge sets,
$\mathcal E_R$ and $\mathcal E_P$, from adjacent residues with consecutive
residue identifiers in the same chain.
Supported insertion-code transitions are included, and an edge is retained
when the C$\alpha$ distance lies between 2.5 and 4.5~\angstrom{}.
Let $m(j)$ denote the predicted site matched to reference residue $j$
under $\mathcal M_3$, and define
\begin{equation}
\mathcal E_R^{\mathrm{rec}}
=
\left\{
(j,k)\in\mathcal E_R:
m(j),m(k)\text{ exist}
\right\}.
\end{equation}
The break rate is
\begin{equation}
\mathrm{Break\ Rate}
=
100\,
\frac{
\sum_{(j,k)\in\mathcal E_R^{\mathrm{rec}}}
\mathbf 1[
(m(j),m(k))\notin\mathcal E_P
\;\land\;
(m(k),m(j))\notin\mathcal E_P
]
}{
|\mathcal E_R^{\mathrm{rec}}|
}.
\label{eq:metric_break_rate}
\end{equation}
Thus, the metric evaluates missing backbone connections only when both
reference endpoints have already been recovered.
Edges with a missing endpoint are excluded from the denominator, and
connectivity in either direction is accepted.

\paragraph{Q-score.}
Q-score measures how well reconstructed protein atoms are supported by the
experimental cryo-EM density \citep{pintilie2020measurement}.
For each atom $b$, the scoring backend samples map density values
$u_{b\ell}$ around the atom and compares them with values
$g_{b\ell}$ from an ideal atom-centered Gaussian profile.
Sampling uses radii from 0 to 2.0~\angstrom{} in 0.1~\angstrom{} increments
with eight sampling positions per radius and Gaussian width
$\sigma=0.6$~\angstrom{}.

For map density $\rho$, the reference profile is
\begin{equation}
g(r)
=
h\exp\left(
-\frac{r^2}{2\sigma^2}
\right)
+b_0,
\end{equation}
where
\begin{equation}
b_0
=
\max(\overline{\rho}-\sigma_\rho,\rho_{\min}),
\qquad
h
=
\min(\overline{\rho}+10\sigma_\rho,\rho_{\max})-b_0.
\end{equation}
The atomic Q-score is
\begin{equation}
Q_b
=
\frac{
\sum_{\ell}
(u_{b\ell}-\overline u_b)
(g_{b\ell}-\overline g_b)
}{
\sqrt{
\sum_{\ell}(u_{b\ell}-\overline u_b)^2
\sum_{\ell}(g_{b\ell}-\overline g_b)^2
+\epsilon
}
},
\label{eq:metric_q_atom}
\end{equation}
with $\epsilon=10^{-6}$.
We report the mean over finite Q-scores of recognized protein atoms:
\begin{equation}
Q
=
\frac{1}{
|\mathcal B_{\mathrm{prot}}^{\mathrm{finite}}|
}
\sum_{b\in\mathcal B_{\mathrm{prot}}^{\mathrm{finite}}}
Q_b.
\label{eq:metric_q}
\end{equation}
This is an atom-level mean rather than a residue-weighted or
C$\alpha$-only average.
The frozen scoring implementation uses random seed 20260908.

\paragraph{Hetero-neighborhood localization.}
We evaluate backbone localization near fixed reference hetero components.
Let $H$ contain the reference heavy atoms from the five evaluated hetero
classes: polymeric nucleic acid, glycan, nucleotide/cofactor, other ligand,
and lipid/detergent.
Waters and ions are excluded.
Predicted and reference backbone sites within 4~\angstrom{} of hetero
atoms are defined separately:
\begin{equation}
I_H
=
\left\{
i:
\min_{\mathbf h\in H}
\|\mathbf p_i-\mathbf h\|_2
\leq4~\text{\AA}
\right\},
\qquad
J_H
=
\left\{
j:
\min_{\mathbf h\in H}
\|\mathbf r_j-\mathbf h\|_2
\leq4~\text{\AA}
\right\}.
\label{eq:metric_near_sets}
\end{equation}
We reuse the global matching $\mathcal M_3$ rather than recomputing
matches within the hetero neighborhoods.
The corresponding metrics are
\begin{align}
\mathrm{Near\ Precision}
&=
100\,
\frac{
|\{(i,j)\in\mathcal M_3:i\in I_H\}|
}{
|I_H|
},
\\
\mathrm{Near\ Recall}
&=
100\,
\frac{
|\{(i,j)\in\mathcal M_3:j\in J_H\}|
}{
|J_H|
}.
\label{eq:metric_near_pr}
\end{align}
Only the indicated endpoint is required to lie in the hetero neighborhood,
so the two numerators need not be identical.
The previously reported near-unmatched fraction is
$100\%-\mathrm{Near\ Precision}$.
Maps with an empty corresponding neighborhood have an undefined score.

\paragraph{Protein--hetero contacts.}
Protein--hetero contacts are defined using the same fixed reference hetero
components.
A protein residue and hetero object form a contact when their minimum
heavy-atom distance is at most 4~\angstrom{}.
Polymeric nucleic acids are represented by chain-level hetero objects,
while the other four hetero classes use individual labeled components.

Let $\mathcal C_R$ denote the reference protein--hetero residue--object
contacts.
Predicted protein residues are mapped to their reference partners through
the backbone matching $\mathcal M_3$, producing mapped predicted contacts
$\widetilde{\mathcal C}_P$.
Unmatched predicted residues retain distinct identifiers outside the
reference residue set, so their contacts remain false positives.
We define
\begin{equation}
\mathrm{TP}
=
|\widetilde{\mathcal C}_P\cap\mathcal C_R|,
\qquad
\mathrm{FP}
=
|\widetilde{\mathcal C}_P\setminus\mathcal C_R|,
\qquad
\mathrm{FN}
=
|\mathcal C_R\setminus\widetilde{\mathcal C}_P|.
\end{equation}
Contact precision and recall are
\begin{equation}
\mathrm{Contact\ Precision}
=
100\,
\frac{\mathrm{TP}}{\mathrm{TP}+\mathrm{FP}},
\qquad
\mathrm{Contact\ Recall}
=
100\,
\frac{\mathrm{TP}}{\mathrm{TP}+\mathrm{FN}}.
\label{eq:metric_contact_pr}
\end{equation}
Contacts are unique protein-residue--hetero-object pairs rather than
counts of individual contacting atom pairs.
Residue matching is geometric and does not require the predicted amino
acid identity to be correct.
Because hetero coordinates are fixed to the reference structure, these
metrics evaluate the reconstructed protein environment around hetero
components rather than hetero detection or hetero structure prediction.


\section{Additional Results}
\label{app:additional_results}

\subsection{Hetero Detection Results}
\label{app:hetero_detection_results}

\begin{table}[t]
\centering
\caption{Class-wise hetero detection performance.
Precision, recall, and F$_1$ are reported for the five hetero classes.}
\label{tab:hetero_detection}
\small
\setlength{\tabcolsep}{7pt}
\renewcommand{\arraystretch}{1.12}
\begin{tabular}{lrrr}
\toprule
Class & Precision (\%) & Recall (\%) & F$_1$ (\%) \\
\midrule
Nucleic acid        & 95.71 & 63.08 & 76.04 \\
Glycan              & 45.33 & 15.20 & 22.77 \\
Nucleotide/cofactor & 80.76 & 16.10 & 26.85 \\
Other ligand        & 51.52 & 46.90 & 49.10 \\
Lipid/detergent     & 15.15 & 13.13 & 14.07 \\
\bottomrule
\end{tabular}
\end{table}

The detector is designed to provide hetero context rather than to reconstruct
complete hetero structures.
Stage~1 uses its multiscale dense features, while Stage~2 uses sparse
candidates whose contributions are controlled by confidence and learned
gates.
Therefore, partial but reliable hetero predictions can still provide useful
information for protein reconstruction.
This setting differs from many component-specific approaches, which
either focus on a single molecular family or assume additional information
such as nucleic-acid sequences, candidate ligand identities, binding sites,
or fitted protein structures
\citep{mostosi2020haruspex,wang2021detecting,li2025all,li2024automated,
sweeney2024chemem,muenks2023automatic,ansell2023lipidens}.

\end{document}